\pdfoutput=1

\documentclass[11pt]{article}

\usepackage[final]{acl}

\usepackage{times}
\usepackage{latexsym}

\usepackage[T1]{fontenc}

\usepackage[utf8]{inputenc}

\usepackage{microtype}

\usepackage{inconsolata}

\usepackage{multirow}
\usepackage{makecell}
\usepackage{hhline}
\usepackage{booktabs}
\usepackage{bm}
\usepackage{amssymb}
\usepackage{amsmath}
\usepackage{url}
\usepackage{graphicx}
\usepackage{float}
\usepackage{subfigure}
\usepackage{subcaption}
\usepackage{paralist}
\usepackage[ruled, lined, linesnumbered, commentsnumbered, longend]{algorithm2e}
\usepackage{xcolor}
\usepackage{colortbl}
\usepackage{pifont}
\usepackage{tabularx}
\usepackage[T2A,LAE,T1]{fontenc}
\usepackage{CJKutf8}
\usepackage{enumitem}
\setlist[itemize]{noitemsep, topsep=0pt}
\usepackage[russian,english]{babel}

\DeclareRobustCommand{\cyrins}[1]{
  \begingroup\fontfamily{cmr}
  \foreignlanguage{russian}{#1}
  \endgroup
}
\DeclareRobustCommand{\aratext}[1]{
  \begingroup\fontfamily{Arab}
  \foreignlanguage{arabic}{#1}
  \endgroup
}

\definecolor{LightCyan}{rgb}{0.88,1,1}
\newcommand{\boldtitle}[1]{\noindent\textbf{#1\ \ }}
\newcommand{\inlinetitle}[1]{\paragraph{#1}}

\newcommand{\ours}{\texttt{CLASS}\xspace}
\newcommand{\oursus}{\texttt{CLASS-US}\xspace}
\newcommand{\oursusstageone}{\texttt{CLASS-US-Stage1}\xspace}
\newcommand{\ourszs}{\texttt{CLASS-ZS}\xspace}

\usepackage{ntheorem}

\theoremstyle{nonumberplain}

\def\eg{{e.g.,}\xspace}
\def\ie{{i.e.,}\xspace}
\def\versus{{\em v.s.}\xspace}

\definecolor{lightblue}{HTML}{bdd6fb}
\definecolor{boxgray}{gray}{0.9}
\definecolor{bgyellow}{HTML}{fcebde}
\definecolor{bgred}{HTML}{d77470}
\definecolor{bggrey}{HTML}{dcc0e5}

\usepackage{inconsolata}
\usepackage{pifont}
\usepackage[most]{tcolorbox}
\usepackage{csquotes}
\usepackage{anyfontsize}
\usepackage{comment}
\usepackage{float}
\usepackage{cuted}
\usepackage{tikz}
\usepackage{listings,multicol}
\usepackage{etoolbox}
\BeforeBeginEnvironment{lstlisting}{\begin{mdframed}\vspace{-0.7em}}
\AfterEndEnvironment{lstlisting}{\vspace{-0.5em}\end{mdframed}}

\def\ifempty#1{\def\temparg{#1}\ifx\temparg\empty}

\usepackage{caption}

\newtcolorbox[list inside=prompt,auto counter,number within=section]{prompt}[1][]{
    colbacktitle=black!60,
    coltitle=white,
    colback=bgyellow,
    fontupper=\footnotesize,
    boxsep=5pt,
    left=0pt,
    right=0pt,
    top=0pt,
    bottom=0pt,
    boxrule=1pt,
    #1,
}

\def\bng{\bngx}

\font\bngx=bang10

\def\*#1*#2{o\null{#2}{#1}}

\title{Pre-training Cross-lingual Open Domain Question Answering \\ with Large-scale Synthetic Supervision}

\author{Fan Jiang \and Tom Drummond \and Trevor Cohn~\Thanks{Now at Google DeepMind} \\
  School of Computing and Information Systems \\
  The University of Melbourne, Victoria, Australia \\
  \texttt{fan.jiang1@student.unimelb.edu.au}\\
  \texttt{\{tom.drummond, trevor.cohn\}@unimelb.edu.au}}

\begin{document}
\maketitle
\begin{abstract}

Cross-lingual open domain question answering (CLQA) is a complex problem, comprising cross-lingual retrieval from a multilingual knowledge base, followed by answer generation in the query language. Both steps are usually tackled by separate models, requiring substantial annotated datasets, and typically auxiliary resources, like machine translation systems to bridge between languages. In this paper, we show that CLQA can be addressed using a single encoder-decoder model. To effectively train this model, we propose a self-supervised method based on exploiting the cross-lingual link structure within Wikipedia. We demonstrate how linked Wikipedia pages can be used to synthesise supervisory signals for cross-lingual retrieval, through a form of cloze query, and generate more natural questions to supervise answer generation. Together, we show our approach, \ours, outperforms comparable methods on both supervised and zero-shot language adaptation settings, including those using machine translation.

\end{abstract}

\section{Introduction}

Open Domain Question Answering (QA) is the task of generating an answer for a given question based on the evidence gathered from a large collection of documents. A widely adopted pipeline "\emph{retrieve-then-read}" is employed for this task~\citep{chen-etal-2017-reading, karpukhin-etal-2020-dense}, which begins by retrieving a small set of passages using a dense retrieval model and subsequently processes retrieved passages to generate the answer with a dedicated reader. Unlike English open-domain QA, where both questions and knowledge sources share the same language, multilingual open-domain QA presents new challenges, as it involves retrieving evidence from multilingual corpora, considering that many languages lack comprehensive support documents or the questions require knowledge from diverse cultures~\citep{asai2021one}.

Several attempts have been made to enhance the performance of multilingual open-domain QA~\citep{asai2021one, lapca}. These approaches typically require passage labels for retriever training through supervised contrastive learning. This requirement complicates cross-lingual retrieval training significantly due to the challenge of constructing a large-scale dataset containing query-passage labels. This challenge emerges from the unavailability of prior knowledge regarding which language contains the relevant evidence. Furthermore, these efforts often involve separate training of the retriever and reader, leading to error propagation within the resulting pipeline.  

Evidence in the context of English open-domain QA reveals that integrating retriever and reader training typically leads to improved performance on both components. This achievement is often realised by training both components~\citep{realm, rag} or a unified model that performs both tasks~\citep{lee-etal-2022-need, jiang-etal-2022-retrieval} through fully end-to-end training. Nonetheless, such a joint training paradigm has not been extensively explored in multilingual open-domain QA, and how to adapt it to suit the complexities of multilingual settings remains an open question.

\begin{figure*}
    \centering
    \includegraphics[width=\linewidth]{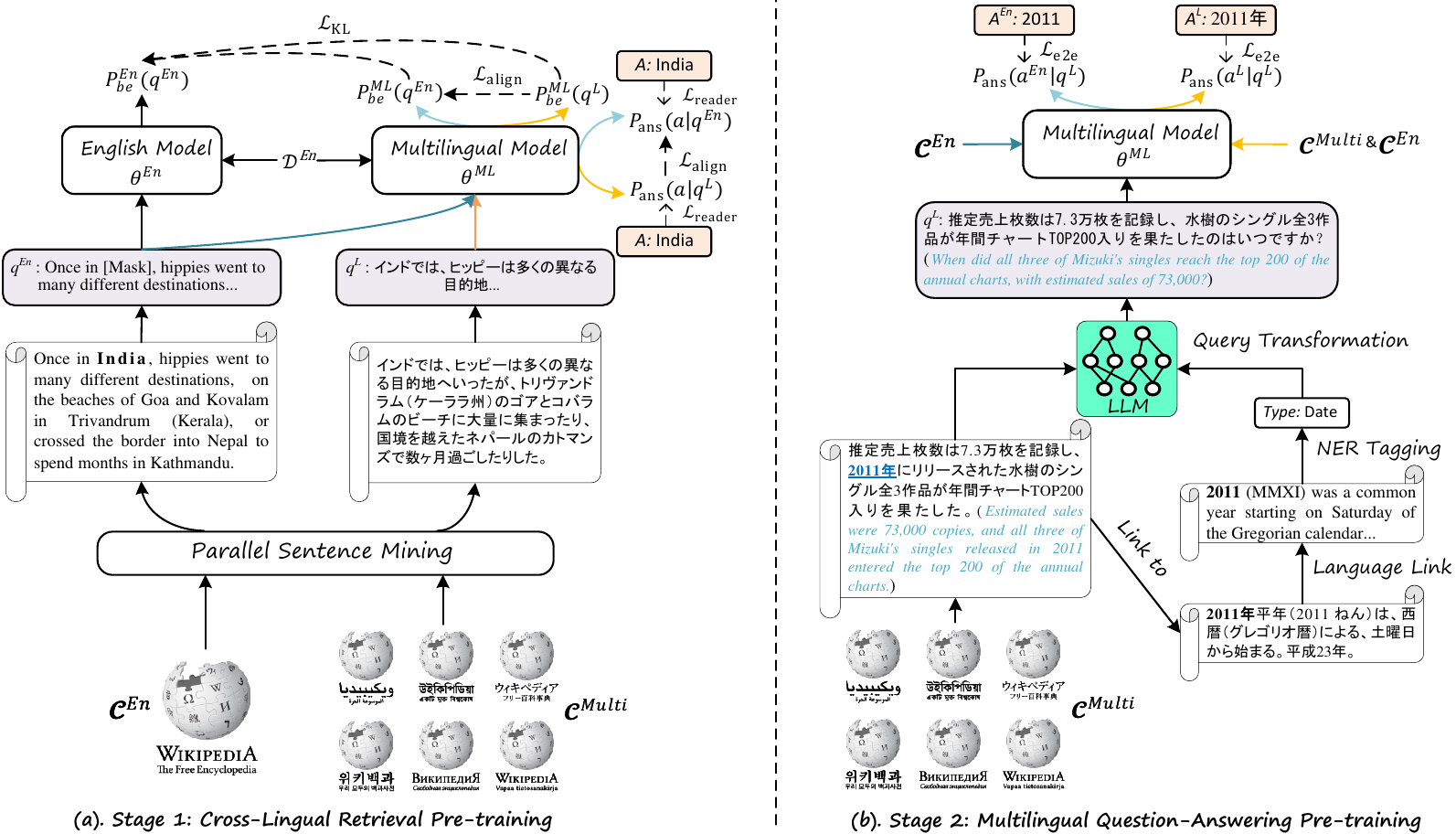}
    \caption{The overview of our two-stage unsupervised pre-training method for cross-lingual open domain question answering. English translations from Google Translate are added in (b) for readability.}
    \label{fig:training_pipeline}
\end{figure*}

In this paper, we introduce the first \emph{unified model} capable of performing both cross-lingual retrieval and multilingual open-domain QA tasks. To achieve this, we propose \textbf{\ours} (\textbf{C}ross-\textbf{L}ingual Q\textbf{A} Pre-training with \textbf{S}ynthetic \textbf{S}upervision), a self-supervised method to pre-train the model with multilingual texts at scale. \ours comprises two core components: \textbf{cross-lingual retrieval pre-training} that equips the model with robust cross-lingual retrieval ability, and \textbf{multilingual QA pre-training} that further enhances retrieval and QA abilities jointly. Concretely, as depicted in Figure~\ref{fig:training_pipeline}, the pre-training data is created by mining parallel queries from parallel Wikipedia pages, using salient entities within English sentences as answers. To facilitate cross-lingual retrievals, a knowledge distillation process is introduced, requiring the model to match the distributions of a well-trained English teacher when given queries in both languages. The follow-up is a self-supervised learning task for end-to-end pre-training by propagating training signals derived from the end QA task. This process entails generating pre-training data using anchor texts indicated by hyperlinks and a \emph{question transformation} technique to resemble the formats of natural questions. Notably, our approach does not necessitate additional tools such as machine translation and offers a more convenient application to low-resource languages, requiring only comparable documents (\ie Wikipedia language links).


This large-scale pre-training framework empowers the model to demonstrate promising unsupervised performance, and it can even outperform many competitive supervised counterparts. By fine-tuning it with supervised English and multilingual QA data, we can attain further improvements, ultimately establishing new state-of-the-art performance in both cross-lingual retrieval and multilingual open-domain QA tasks. In summary, our contributions are:\footnote{Code and data are available \href{https://github.com/Fantabulous-J/CLASS}{here}.}


\begin{compactenum}
    \item Empirical results on the \textsc{Xor-TyDi QA} benchmark demonstrate that \ours outperforms a wide range of prominent unsupervised, zero-shot, and supervised models on both tasks, while solely relying on QA pairs throughout the whole training processes.
    \item On the MKQA dataset, \ours exhibits remarkable generalisation capabilities across linguistically diverse languages without using human-annotated data.
    \item To the best of our knowledge, we are the pioneers in systematically exploring the advantages of pre-training for multilingual retrieval and open-domain QA tasks. This demonstrates the feasibility of achieving multilingual open-domain QA within a unified model.
\end{compactenum}

\section{Preliminaries}

\subsection{Task Definition}

Given a query $q^L$ in language $L$, \textbf{Cross-lingual Passage Retrieval} requires retrieving a collection of passages $\mathcal{D}^{En}$ from English Wikipedia $\bm{C^{\textbf{En}}}$ that potentially provide evidence to answer $q^L$. In contrast, \textbf{Multilingual Open-Domain Question Answering} aims at answering $q^L$ in language $L$ by referring to a multilingual Wikipedia $\bm{C^{\textbf{Multi}}}$. In this setting, the prior knowledge of which language contains the evidence is unavailable, and the relevant passages can be retrieved from any language.

\begin{figure}
    \centering
    \includegraphics[width=\linewidth]{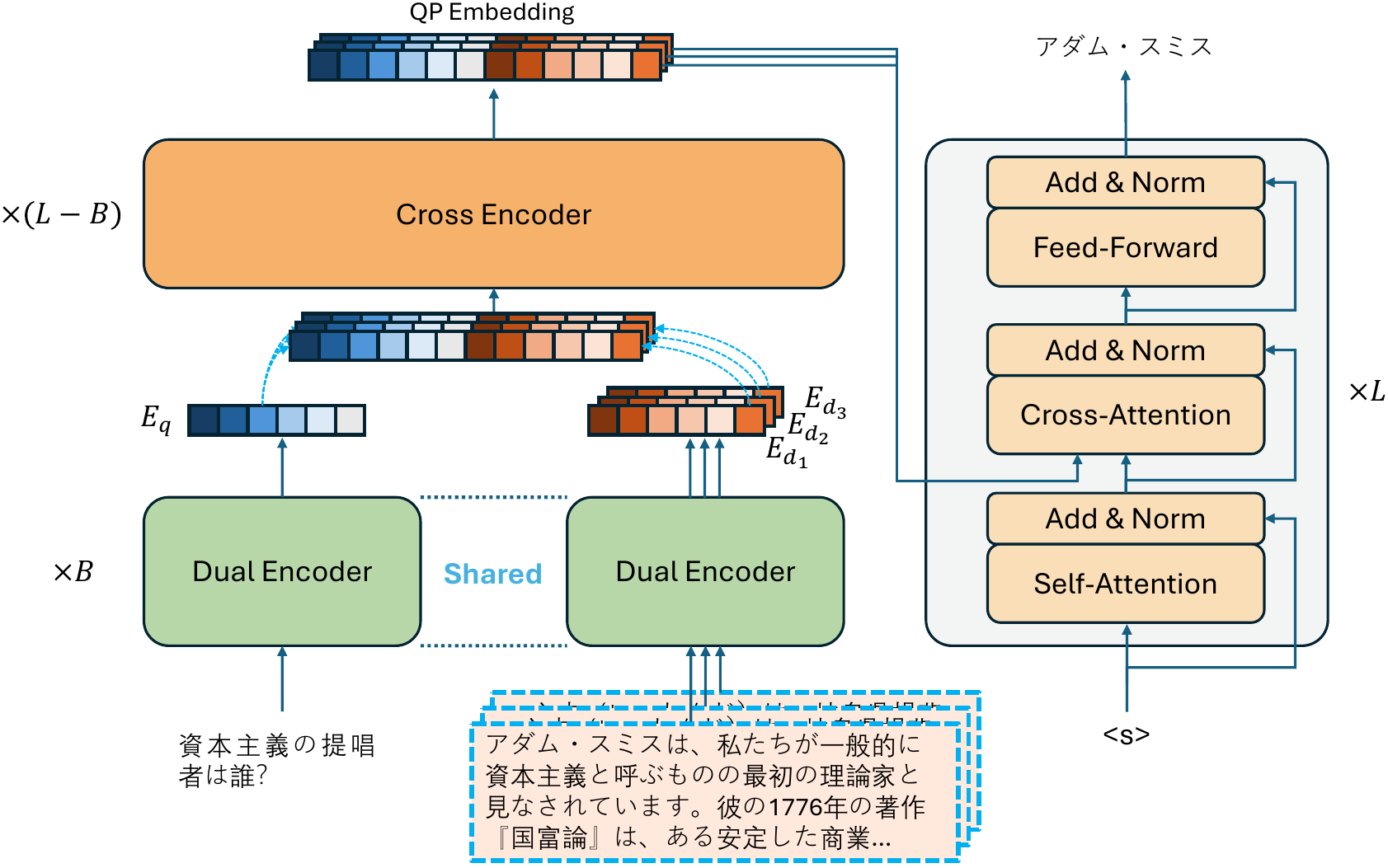}
    \caption{The unified model for passage retrieval and question answering.}
    \label{fig:unified model}
\end{figure}

\subsection{Model Architecture}

Figure~\ref{fig:unified model} shows the overall structure of our model. In this model, the bottom layers of the encoder function as the \emph{retriever}, encoding queries and passages independently for efficient retrieval. The remaining encoder layers and the entire decoder are designated as the \emph{reader} for question answering.

\inlinetitle{Retriever.} 
The retriever is a bi-encoder that uses the first $B$ encoder layers with $H$ heads to encode query $q$ and passages $d$ from a corpus $\mathcal{D}$. We use the query $Q$ and key vectors $K$ in $B+1$-th layer as their embeddings, respectively~\citep{jiang-etal-2022-retrieval}:
\begin{align}
    & \scalebox{0.9}{$E_{\mathbf{d}}=\{K_{\mathbf{d}}^{B+1, h} \in \mathbb{R}^{|\mathbf{d}| \times e}\}_{h=1}^H, \nonumber$} \\
    & \scalebox{0.9}{$E_{\mathbf{q}}=\{Q_{\mathbf{q}}^{B+1, h} \in \mathbb{R}^{|\mathbf{q}| \times e}\}_{h=1}^H, \nonumber$}
\end{align}
where $|\mathbf{q}|$ and $|\mathbf{d}|$ are sequences lengths and $e$ is the dimension of each head.

The self-attention matrix $\operatorname{SA}_{\mathbf{q},\mathbf{d}}^{B+1, h}$ from a specific head ($h=6$~\citep{jiang-etal-2022-retrieval}) is considered the source of retrieval scores. A sum of max computations~\citep{colbertv1} is performed to reduce it to yield the retrieval score:
\begin{align}
   & \scalebox{0.9}{$s_{\text{mv}}(q,d) = \sum_{i\in|\mathbf{q}|}\max_{j\in|\mathbf{d}|}\operatorname{SA}_{i,j}^{B+1, h}, \nonumber$} \\
   & \scalebox{0.9}{$\operatorname{SA}_{\mathbf{q},\mathbf{d}}^{B+1, h} = Q_{\mathbf{q}}^{B+1, h}\times K_{\mathbf{d}}^{B+1, h^\top} \in \mathbb{R}^{|\mathbf{q}| \times |\mathbf{d}|}. \nonumber$}
\end{align}
We denote this as \textbf{Multi-Vector Retrieval} and consider it as our \emph{default setting}. We also explore \textbf{Dense Retrieval}, which takes the average pooling of layer $B$'s output with $\operatorname{LayerNorm}$ as query $Q_{\mathbf{q}}$ and passage $K_{\mathbf{d}}$ representations, and the relevance is measured by their dot product:
\begin{equation}
    \scalebox{0.9}{$s_{\text{dense}}(q,d)=\operatorname{LN}(Q_{\mathbf{q}})\cdot\operatorname{LN}(K_{\mathbf{d}}). \nonumber$}
\end{equation}
The top-$k$ most relevant passages are then retrieved by $\mathcal{D}_q=\arg\operatorname{topk}_{d_i\in\mathcal{D}}P_{\text{be}}(\cdot|q,D)=\arg\operatorname{topk}\left[s(q,d_0), \dots, s(q,d_{|{d\in\mathcal{D}}|})\right]$.

\inlinetitle{Reader.}


The encoded query and each top-$k$ passage in $\mathcal{D}_q$ are concatenated and fed into the remaining \emph{cross-encoder} layers. Finally, the joint encodings $\{E_{\mathbf{q,d_i}}\}_{i=0}^{|\mathcal{D}_q|}$ are integrated into the decoder through cross-attention to generate the answer $a$ efficiently~\citep{izacard-grave-2021-leveraging}: $P_{\text{ans}}(a|q,\mathcal{D}_q)=\log \prod_{t=1}^T P(a_t|a_{<t},q,\mathcal{D}_q)$.
\section{Method}
We propose an unsupervised two-stage pre-training method for cross-lingual open-retrieval question answering, as depicted in Figure~\ref{fig:training_pipeline}. Our approach starts with \textbf{cross-lingual retrieval pre-training}, where the \emph{unified multilingual model} develops excellent cross-lingual dense retrieval capabilities. This proficiency is acquired through learning from a well-trained English model, employing cloze-style parallel queries and retrieved English passages as inputs. The subsequent stage involves \textbf{pre-training for multilingual question-answering (QA)}, where the \emph{unified model} is further pre-trained on multilingual question-answer pairs that are automatically generated. This process entails selecting potential answers from anchor texts and applying our novel \emph{question transformation} techniques to convert cloze questions into natural questions by prompting a large language model.

\subsection{Cross-Lingual Retrieval Pre-training}\label{sec:stage1_pretrain}

\inlinetitle{Pre-training Data.}
We consider cloze questions, which are statements with the answer masked, as pseudo queries. The answers are salient spans selected from named entities. We extract all named entities for an English sentence using a NER system, generating queries for each. Formally, let $s^{En}$ be a sentence sampled from an English Wikipedia page $\mathcal{W}^{En}$, along with its associated named entities $\{a_i\}^n_{i=1}$. This allows us to derive cloze queries $\{q_i^{En}\}^n_{i=1}$ by masking each entity $a_i$. Then, for each $q_i^{En}$, the objective is to identify its translation $q_i^L$ in language $L$ by searching from sentences $\{q_j^L\}_{j=0}^n$ within a Wikipedia page $\mathcal{W}^L$, which is connected to $\mathcal{W}^{En}$ via language links in Wikipedia. 

We use a margin-based mining method~\citep{artetxe-schwenk-2019-margin} to identify parallel sentences based on their similarity in the embedding space:
\begin{equation}
    \scalebox{1}{$\operatorname{M}(q_i,q_j) = \frac{\cos (q_i,q_j)}{\sum_{z\in N_{q_i}}\frac{\cos (q_i,z)}{2k}+\sum_{z\in N_{q_j}}\frac{\cos (q_j,z)}{2k}}, \nonumber$}
\end{equation}
where $N_{q_i}$ and $N_{q_j}$ are the top-$k$ neighbours of sentence $q_i$ and $q_j$ in the other language, respectively. $cos(q_i,q_j)$ denotes the cosine similarity between the embeddings of $q_i$ and $q_j$ extracted using \texttt{mSimCSE}~\citep{wang-etal-2022-english}. 
We apply this scoring function to $q_i^{En}$ and each $q_j^L\in\{q_j^L\}_{j=0}^n$. Pairs whose scores surpass a threshold $T$ are selected as parallel queries, denoted as $\{q_i^{En}, q_j^L, a_i\}$.\footnote{We identify $a_i$ and mask it in $q_j^L$ through string match if $L$ is written in Latin script and leave $q_j^L$ unchanged otherwise.}

\inlinetitle{Training.}
A well-trained English model $\theta^{En}$ is employed to teach a multilingual model $\theta^{ML}$ using parallel queries. Specifically, given a training example $\{q^{En}, q^L, a\}$, we employ $\theta^{En}$ to retrieve a set of relevant passages $\mathcal{D}_{q^{En}}$ from English Wikipedia $\bm{C^{\textbf{En}}}$ for $q^{En}$. The multilingual model is then compelled to align its retrieval distributions with those of $\theta^{En}$ over $\mathcal{D}_{q^{En}}$ through KL divergence loss:
\begin{align}
    \mathcal{L}_{\operatorname{KL}} & \resizebox{0.435\textwidth}{!}{$ = \mathbb{KL}(P^{ML}_{\text{be}}(\cdot|q^L,\mathcal{D}_{q^{En}})||P^{En}_{\text{be}}(\cdot|q^{En},\mathcal{D}_{q^{En}})) \nonumber$} \\
    \quad & \resizebox{0.435\textwidth}{!}{$ + \mathbb{KL}(P^{ML}_{\text{be}}(\cdot|q^{En},\mathcal{D}_{q^{En}})||P^{En}_{\text{be}}(\cdot|q^{En},\mathcal{D}_{q^{En}})). \nonumber$}
\end{align}

Additionally, $\theta^{ML}$ is trained to predict the answer $a$ with either $q^{En}$ or $q^L$ as the question:
\begin{equation}
    \scalebox{0.95}{$\mathcal{L}_{\text{reader}} = -P_{\text{ans}}(a|q^{En},\mathcal{D}_{q^{En}}) - P_{\text{ans}}(a|q^L,\mathcal{D}_{q^{En}}). \nonumber$}
\end{equation}

Moreover, to ensure that the multilingual model generates consistent predictions across languages, we introduce an alignment regularisation term:
\begin{align}
    \mathcal{L}_{\text{align}} & \resizebox{0.425\textwidth}{!}{$= \mathbb{KL}(P^{ML}_{\text{be}} (\cdot|q^L,\mathcal{D}_{q^{En}})||P^{ML}_{\text{be}}(\cdot|q^{En},\mathcal{D}_{q^{En}})) \nonumber$} \\
    \quad & \resizebox{0.423\textwidth}{!}{$ + \mathbb{KL}(P_{\text{ans}}(a|q^L,\mathcal{D}_{q^{En}})||P_{\text{ans}}(a|q^{En},\mathcal{D}_{q^{En}})). \nonumber$}
\end{align}

Overall, $\theta^{ML}$ is trained with the weighted combined loss: $\mathcal{L}_{\text{stage1}}=\mathcal{L}_{\text{reader}}+\alpha\cdot(\mathcal{L}_{\operatorname{KL}}+\mathcal{L}_{\text{align}})$.

\subsection{Multilingual QA Pre-training}\label{sec:stage2_pretrain}
The cloze questions used in \S\ref{sec:stage1_pretrain} are substantially different from the formats of natural questions asked by real users, which inherently impedes the development of advanced QA skills. Moreover, the incapacity to precisely locate and mask the answer $a$ within $q^L$ for perfectly aligned queries makes the QA task notably simpler, as $a$ implicitly appears in $q^L$ (\eg "\begin{CJK}{UTF8}{min}インド\end{CJK}" in $q^L$ is the Japanese answer in Figure~\ref{fig:training_pipeline} (a)). Meanwhile, since $q^{En}$ and $q^L$ could be roughly aligned, the querying of $a$ by $q^L$ is not assured, thereby introducing noise into the pre-trained data (\eg "In \emph{1945}, his father sent him to Collège des Frères" and "\begin{CJK}{UTF8}{min}父はサブリーをヤッファのカトリック系フランス語学校に送った。\end{CJK}" are aligned but the Japanese query does not mention the answer \emph{1945}). Thus, we design another pre-training technique to address the limitations above.

\subsubsection{Pre-training Data}\label{sec:stage2_pretrain_data}
The construction of pre-training data in this stage involves two sequential steps. Initial data are first acquired from a multilingual Wikipedia source in the format of cloze questions, followed by a format transformation into natural questions.

\inlinetitle{Initial Data.}
In contrast to English texts, where robust NER systems facilitate the detection of named entities with high precision for answer generation, such systems in other languages exhibit inherent deficiencies. Instead, we employ anchor texts with hyperlinks as answer candidates. Specifically, for a given sentence $s^L$ in language $L$, we consider the anchor texts $\{a_i^L\}_{i=0}^n$ within it as potential answers and construct cloze questions $\{s_i^L\}_{i=0}^n$ accordingly. 

For each $a_i^L$, we fetch the Wikipedia page $\mathcal{W}^L$ to which it links and access the corresponding English Wikipedia page $\mathcal{W}^{En}$ via language link. Subsequently, the title $a_i^{En}$ of $\mathcal{W}^{En}$ is assumed to be the pseudo translation of $a_i^L$ (Figure~\ref{fig:training_pipeline} (b)). Moreover, NER tagging is performed on the first paragraph of $\mathcal{W}^{En}$ to identify the type $t_i$ of the title entity $a_i^{En}$, which is then assigned to $a_i^L$. Finally, a training example is derived as $(s_i^L, a_i^L, a_i^{En}, t_i)$.

\inlinetitle{Query Transformation.}
We employ large language models (LLMs) for query transformation via In-Context Learning (ICL)~\citep{gpt3-few-shot}. 

We first prompt ChatGPT (\texttt{gpt-3.5-turbo}) to generate a few examples as meta-examples~\citep{fan2023improving} for ICL. Specifically, we randomly sample instances from the initial dataset and generate transformed questions based on the structure of the prompt shown in Prompt~\ref{prompt:meta_example}.

\begin{prompt}[title={Prompt \thetcbcounter: Meta-Example Generation}, label=prompt:meta_example]
Rewrite this sentence $\{\textcolor{red}{s_i^L}\}$ into a natural question whose question word is $\{\textcolor{red}{\texttt{wh\_word}}\}$ and answer is $\{\textcolor{red}{a_i^L}\}$. Please respond in the format: "The transformed question is: $\{\textcolor{red}{q_i^L}\}$"
\end{prompt}

where \textcolor{red}{\texttt{wh\_word}} is chosen according to the entity type $t_i$ through heuristics~\citep{lewis-etal-2019-unsupervised}. This step yields a curated set of ICL examples: $\mathbb{K}=\{c_i^L, \texttt{wh\_word}, a_i^L, q_i^L\}_{i=0}^k$. An example is shown in Figure~\ref{fig:chatgpt_prompt_example} in the Appendix.


Subsequently, the curated ChatGPT examples are used as the source to few-shot prompt a smaller LLM, LLaMA-2-7B~\citep{touvron2023llama}, to generate many more instances efficiently. We include the prompting examples in Appendix~\ref{appendix:icl_examples}.

\subsubsection{Joint Training}
The retriever learns indirectly from the answer generation task, taking the cross-attention score from the decoder as the target for query-passage relevance measurement~\citep{izacard2021distilling}:
\begin{gather}
    \scalebox{0.9}{$\mathcal{L}_\text{KL} = \mathbb{KL}(P_{\text{be}}(\cdot|q^L,\mathcal{D}_{q^L})||P_{\text{ca}}(\cdot|q^L,\mathcal{D}_{q^L})),\nonumber$} \\
    \scalebox{0.9}{$P_{\text{ca}}(d_i|q^L,\mathcal{D}_{q^L}) = \sum_{h=0}^H\sum_{t=0}^{|d_i|}\frac{\operatorname{SG}(\operatorname{CA}(0,h,t))}{H}\ |\ d_i\in\mathcal{D}_{q^L}, \nonumber$}
\end{gather}
where $\mathcal{D}_{q^L}$ is the set of passages returned by the retriever itself and $P_{\text{ca}}$ is the target distribution gathered from the decoder's cross-attention scores. $\operatorname{SG}$ signifies stop-gradient, which blocks the gradient to ensure the decoder is not affected by the retriever loss, and $\operatorname{CA}$ denotes the cross-attention score at the last decoder layer. The term 0 refers to the first output token, $H$ is the number of cross-attention heads, and $|d_i|$ is the length of passage $d_i$.

The reader optimises the negative log-likelihood of generating $a^L$ given $q^L$ and relevant passages $\mathcal{D}_{q^L}$ as input: $\mathcal{L}_\text{reader}=-P_{\text{ans}}(a^L|q^L,\mathcal{D}_{q^L})$.
The final loss combines reader and retriever loss: $\mathcal{L}_{\text{e2e}}=\alpha\cdot\mathcal{L}_\text{KL}+\mathcal{L}_\text{reader}$.

\inlinetitle{Asynchronous Passage Update.}
During training, we need to use the retriever to gather a set of passages $\mathcal{D}_{q^L}$ from $\bm{C^{\textbf{Multi}}}$ for each $(q^L, a^L)$.\footnote{We replace $a^L$ with $a_i^{En}$ and $\bm{C^{\textbf{Multi}}}$ with $\bm{C^{\textbf{En}}}$ when focusing on cross-lingual retrieval from English corpus.} However, since the retriever's parameters are updated constantly, employing the latest model for retrieval becomes computationally expensive due to the need for recomputing all passage embeddings. To ensure efficient training, 
we periodically update the retrieved passages for each training query using the
most recent model every 1000 steps.
\section{Experiments}



\inlinetitle{Datasets, Baselines and Metrics.}
We evaluate our model on the \textsc{Xor-TyDi QA} dataset~\citep{asai-etal-2021-xor}, with XOR-Retrieve for cross-lingual retrieval, and XOR-Full for multilingual open-domain QA. We employ MKQA~\citep{longpre-etal-2021-mkqa} for zero-shot evaluation on unseen languages. We use the February 2019 English Wikipedia dump as $\bm{C^{\textbf{En}}}$ and use the Wikipedia dumps of the same date, consisting of 13 diverse languages from all 7 languages of \textsc{Xor-TyDi QA} and a subset of MKQA languages as $\bm{C^{\textbf{Multi}}}$~\citep{asai-etal-2021-xor}.

We compare retrieval performance with translate-test methods DPR+MT~\citep{asai-etal-2021-xor}, multilingual dense passage retrievers mDPR, CORA, Sentri, QuiCK, LAPCA, SWIM-X~\citep{asai-etal-2021-xor, asai2021one, sorokin-etal-2022-ask, ren-etal-2022-empowering, lapca, n2023leveraging}, and multi-vector retriever DrDecr~\citep{li-etal-2022-learning-cross}. We report top-$n$ retrieval accuracy, the fraction of queries for which the top-$n$ retrieved tokens contain the answer. We compare QA results with multilingual models that use BM25 for monolingual retrieval, translate-test models MT+DPR, GMT+GS, MT+Mono and ReAtt+MT~\citep{asai-etal-2021-xor, jiang-etal-2022-retrieval}, and multilingual fusion-in-decoder models CORA, Sentri and LAPCA using F1, exact match (EM) and BLEU scores.

\subsection{Experimental Settings}
\inlinetitle{Pre-training Corpus.}
In cross-lingual retrieval pre-training, we gather the parallel pages across various languages for each $\mathcal{W}^{En}\in\bm{C^{\textbf{En}}}$. We consider 15 distinct languages, with 7 from \textsc{Xor-TyDi QA} and 8 being high-resource or closely related to the 7 evaluated languages. Parallel sentences are mined from each pair of parallel pages. A state-of-the-art NER tagger is applied to each English sentence, and we retain pairs that contain named entities. 

In multilingual QA pre-training, data generation is limited to 7 languages on \textsc{Xor-TyDi QA}. We employ LLaMA-2-7B to generate one transformed question per training example with 3 randomly sampled meta-examples in the same language as the prompt. We generate multiple questions for each example in low-resource languages.
More details are in Appendices~\ref{appendix:parallel_mining} and \ref{appendix:query_transform}.

\begin{figure}[t]
    \centering
    \includegraphics[width=\linewidth]{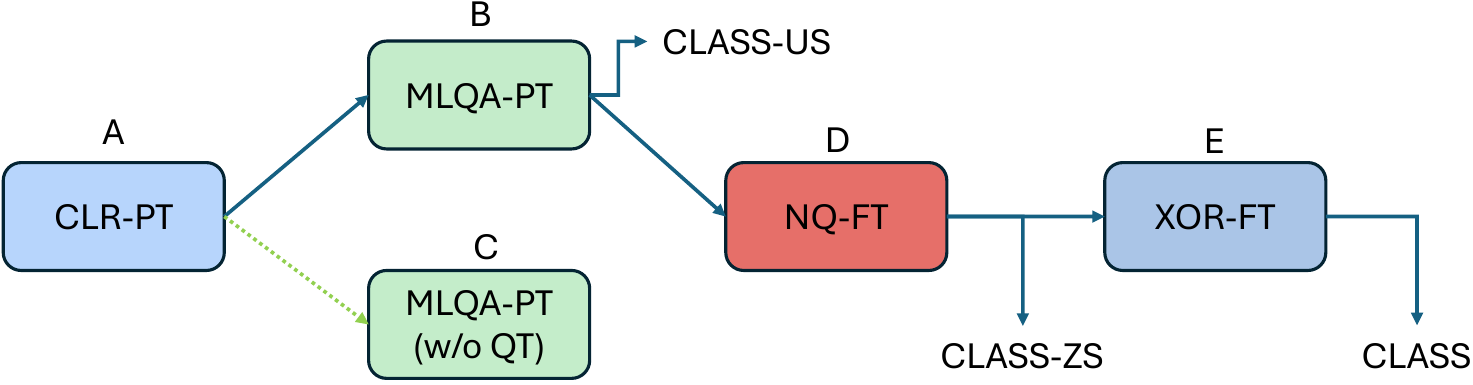}
    \caption{Our training pipeline. CLR: cross-lingual retrieval, MLQA: multilingual question answering, QT: query transformation, PT: pre-training, FT: fine-tuning.}
    \label{fig:training_recipe}
\end{figure}

\begin{table*}[t]
\setlength{\belowcaptionskip}{-0.2cm}
\setlength{\tabcolsep}{3.5pt}
\centering
\resizebox{\linewidth}{!}{\begin{tabular}{l|ccc|cccccccc|cccccccc}
    \toprule
    \multirow{2}{*}[-1ex]{\bf Method} & \multirow{2}{*}[-1ex]{\shortstack{\bf \# Total \\ \bf Params}} & \multirow{2}{*}[-1ex]{\shortstack{\bf Pre-train \\ \bf Data}} & \multirow{2}{*}[-1ex]{\shortstack{\bf Fine-tuning \\ \bf Data}} & \multicolumn{8}{c|}{\bf R@2kt} & \multicolumn{8}{c}{\bf R@5kt} \\
    \cmidrule(lr){5-12} \cmidrule(lr){13-20}
    & & & & Ar & Bn & Fi & Ja & Ko & Ru & Te & Avg & Ar & Bn & Fi & Ja & Ko & Ru & Te & Avg \\
    \midrule
    \rowcolor{blue!15} \multicolumn{20}{c}{\bf\emph{Unsupervised Retrievers}} \\
    LAPCA$^\S$ & 560M & Wikipedia & --- & 51.1 & 50.2 & 48.6 & 35.1 & \underline{57.3} & 32.2 & 64.4 & 48.4 & 61.0 & 58.4 & 52.6 & 40.5 & \underline{66.7} & 40.8 & 70.1 & 55.7 \\
    SWIM-X & 580M & mC4 & \footnotesize SWIM-IR & 50.8 & 65.1 & 56.1 & \underline{48.1} & \underline{54.0} & 55.7 & \underline{66.4} & 56.6 & 57.9 & \underline{75.0} & \underline{65.6} & \underline{59.3} & 58.9 & 64.6 & \underline{74.4} & 65.1 \\
    \bf \oursus & 410M & Wikipedia & --- & \bf 66.0 & \bf 75.7 & \bf 63.4 & \bf 57.7 & \bf 63.5 & \bf 68.8 & \bf 70.6 & \bf 66.5 & \bf 71.2 & \bf 81.6 & \bf 69.4 & \bf 66.8 & \bf 70.5 & \bf 75.1 & \bf 77.3 & \bf 73.1 \\
    \ \ w/ Dense & 410M & Wikipedia & --- & \underline{54.4} & \underline{67.4} & \underline{58.6} & 47.7 & 51.6 & \underline{59.9} & \underline{65.6} & \underline{57.9} & \underline{64.8} & 73.0 & 64.7 & 57.3 & 58.6 & \underline{67.9} & 70.6 & \underline{65.3} \\
    \rowcolor{blue!15} \multicolumn{20}{c}{\bf\emph{Zero-shot Retrievers}} \\
    DPR+MT$^{\dagger}$ & 220M & --- & \footnotesize NQ & 43.4 & 53.9 & 55.1 & 40.2 & 50.5 & 30.8 & 20.2 & 42.0 & 52.4 & 62.8 & 61.8 & 48.1 & 58.6 & 37.8 & 32.4 & 50.6 \\
    LAPCA$^\S$ & 560M & Wikipedia & \footnotesize NQ+XPAQ & 46.2 & 50.3 & 56.6 & 41.4 & 48.7 & 52.3 & 54.6 & 50.0 & 53.0 & 60.5 & 66.2 & 49.7 & 56.1 & 60.7 & 63.8 & 58.6 \\
    ReAtt+MT & 583M & --- & \footnotesize NQ & \underline{63.1} & 67.7 & 20.7 & \underline{55.9} & \underline{60.3} & \underline{55.3} & 58.4 & 54.5 & \underline{67.3} & 71.0 & 29.3 & \underline{61.8} & \underline{67.0} & \underline{61.2} & 66.4 & 60.6 \\
    \bf \ourszs & 410M & Wikipedia & \footnotesize NQ & \bf 65.1 & \bf 79.3 & \bf 67.8 & \bf 60.6 & \bf 61.1 & \bf 69.2 & \bf 74.4 & \bf 68.2 & \bf 72.5 & \bf 83.2 & \bf 73.9 & \bf 70.5 & \bf 69.1 & \bf 75.1 & \bf 81.9 & \bf 75.2 \\
    \ \ w/ Dense & 410M & Wikipedia & \footnotesize NQ & 59.2 & \underline{70.1} & \underline{59.9} & 51.5 & 57.2 & 51.5 & \underline{72.3} & \underline{60.2} & 66.7 & \underline{78.6} & \underline{66.6} & 60.2 & 63.2 & 58.2 & \underline{78.2} & \underline{67.4} \\
    \rowcolor{blue!15} \multicolumn{20}{c}{\bf\emph{(Semi-) Supervised Retrievers}} \\
    CORA & 557M & --- & \footnotesize NQ+XOR & 32.0 & 42.8 & 39.5 & 24.9 & 33.3 & 31.2 & 30.7 & 33.5 & 42.7 & 52.0 & 49.0 & 32.8 & 43.5 & 39.2 & 41.6 & 43.0 \\
    mDPR$^{\dagger}$ & 557M & --- & \footnotesize NQ+XOR & 38.8 & 48.4 & 52.5 & 26.6 & 44.2 & 33.3 & 39.9 & 40.5 & 48.9 & 60.2 & 59.2 & 34.9 & 49.8 & 43.0 & 55.5 & 50.2 \\
    Sentri$^\S$ & 560M & --- & \footnotesize NQ+TQA+XOR & 47.6 & 48.1 & 53.1 & 46.6 & 49.6 & 44.3 & 67.9 & 51.0 & 56.8 & 62.2 & 65.5 & 53.2 & 55.5 & 52.3 & 80.3 & 60.8 \\
    QuiCK & 557M & --- & \footnotesize NQ+XOR & 52.8 & 70.1 & 62.2 & 54.8 & 62.8 & 57.8 & 70.6 & 61.3 & 63.8 & 78.0 & 65.3 & 63.5 & 69.8 & 67.1 & 74.8 & 68.9 \\
    DrDecr$^{\ast}$ & 278M & WikiMatrix & \footnotesize NQ+XOR & - & - & - & - & - & - & - & 66.0 & 70.2 & 85.9 & 69.4 & 65.1 & 68.8 & 68.8 & 83.2 & 73.1 \\
    LAPCA$^\S$ & 560M & Wikipedia & \footnotesize NQ+XPAQ+XOR & 61.1 & 76.9 & \bf 72.6 & \underline{60.9} & \underline{69.1} & \underline{69.1} & 75.6 & \underline{69.3} & 70.2 & 83.8 & \bf 79.6 & \underline{69.7} & \underline{73.6} & \underline{75.5} & \underline{83.1} & \underline{76.5} \\
    \bf \ours & 410M & Wikipedia & \footnotesize NQ+XOR & \bf 67.3 & \bf 80.9 & \underline{67.2} & \bf 64.7 & \bf 71.6 & \bf 69.6 & \bf 79.8 & \bf 71.6 & \bf 74.8 & \bf 84.5 & \underline{72.3} & \bf 73.9 & \bf 79.3 & \bf 77.2 & \bf 85.3 & \bf 78.2 \\
    \ \ w/ Dense & 410M & Wikipedia & \footnotesize NQ+XOR & \underline{66.7} & \underline{79.6} & 64.3 & 58.1 & 66.0 & 64.1 & \underline{77.7} & 68.1 & \underline{70.6} & \underline{84.9} & 71.0 & 66.0 & 72.6 & 70.0 & 81.9 & 73.9 \\
    \bottomrule
    \end{tabular}}
    \caption{Results on the dev set of XOR-Retrieve. The best and second-best results are marked in \textbf{bold} and \underline{underlined}. $\dagger$ denotes results reported by~\citet{asai-etal-2021-xor}. $\ast$ indicates human-translated supervised parallel queries released by XOR-Retrieve are used for training. $\S$ represents methods that employ MT systems for training data augmentation.}
\label{tab:xor_retrieve_results}
\end{table*}
\begin{table*}[t]
\setlength{\tabcolsep}{6.5pt}
\centering
\resizebox{\linewidth}{!}{\begin{tabular}{l@{\hspace{1cm}}|ccc|ccccccc|ccc}
    \toprule
    \multirow{2}{*}[-1ex]{\bf Method} & \multirow{2}{*}[-1ex]{\shortstack{\bf \# Total \\ \bf Params}} & \multirow{2}{*}[-1ex]{\shortstack{\bf Pre-training \\ \bf Data}} & \multirow{2}{*}[-1ex]{\shortstack{\bf Fine-tuning \\ \bf Data}} & \multicolumn{7}{c|}{F1} & \multicolumn{3}{c}{Macro Average} \\
    \cmidrule(lr){5-11} \cmidrule(lr){12-14}
    & & & & \bf Ar & \bf Bn & \bf Fi & \bf Ja & \bf Ko & \bf Ru & \bf Te & \bf F1 & \bf EM & \bf BLEU \\
    \midrule
    BM25$^\dagger$ & --- & --- & \footnotesize XOR & 31.1 & 21.9 & 21.4 & 12.4 & 12.1 & 17.7 & – & – & – & – \\
    MT+DPR$^\dagger$ & --- & --- & \footnotesize NQ & 7.2 & 4.3 & 17.0 & 7.9 & 7.1 & 13.6 & 0.5 & 8.2 & 3.8 & 6.8 \\
    ReAtt+MT & 1.19B & --- & \footnotesize NQ & 15.0 & 10.5 & 1.8 & 13.1 & 14.9 & 15.4 & 8.2 & 11.3 & 5.5 & 9.5 \\
    GMT+GS$^\dagger$ & --- & --- & \footnotesize NQ & 18.0 & 29.1 & 13.8 & 5.7 & 15.2 & 14.9 & 15.6 & 16.0 & 9.9 & 14.9 \\
    MT+Mono$^\dagger$ & --- & --- & \footnotesize NQ+XOR & 15.8 & 9.6 & 20.5 & 12.2 & 11.4 & 16.0 & 0.5 & 17.3 & 7.5 & 10.7 \\
    CORA$^\dagger$ & 1.14B & --- & \footnotesize NQ+XOR & 42.9 & 26.9 & 41.4 & 36.8 & 30.4 & 33.8 & 30.9 & 34.7 & 25.8 & 23.3 \\
    \bf \ours & 1.23B & Wikipedia & \footnotesize NQ+XOR & \bf 49.5 & \bf 32.0 & \bf 49.6 & \bf 44.7 & \underline{37.5} & \bf 41.4 & \bf 42.0 & \bf 42.4 & \bf 32.7 & \bf 29.2 \\
    \ \ w/ Dense & 1.23B & Wikipedia & \footnotesize NQ+XOR & \underline{49.1} & \bf 32.0 & \underline{46.7} & \underline{44.1} & \bf 38.4 & \underline{39.9} & \underline{41.1} & \underline{41.6} & \underline{32.5} & \underline{28.2} \\
    \rowcolor{blue!15} \multicolumn{14}{c}{\bf\emph{Incomparable Models (for Reference)}} \\
    Sentri$^\S$ & 1.14B & --- & \footnotesize NQ+TQA+XOR & 52.5 & 31.2 & 45.5 & \bf 44.9 & 43.1 & 41.2 & 30.7 & 41.3 & 34.9 & 30.7 \\
    LAPCA$^\S$ & 1.14B & Wikipedia & \footnotesize NQ+XPAQ+XOR & \bf 53.4 & \bf 50.2 & \bf 49.3 & 44.7 & \bf 49.5 & \bf 49.3 & \bf 38.9 & \bf 47.8 & \bf 38.7 & \bf 35.5 \\
    \bottomrule
    \end{tabular}}
    \caption{QA results on the XOR-Full dev set. The best and second-best results are marked in \textbf{bold} and \underline{underlined}. $\dagger$ denotes results from ~\citet{asai2021one}. $\S$ indicates methods that use synthetic and translated English datasets.}
\label{tab:xor_full_reader_results}
\end{table*}

\inlinetitle{Training Sequence.}
Figure~\ref{fig:training_recipe} shows the complete pre-training and fine-tuning sequence. \textit{i) Cross-lingual Retrieval Pre-training} (CLR-PT): We pre-train \texttt{mt5-large}~\citep{xue-etal-2021-mt5} as in \S\ref{sec:stage1_pretrain} to get \oursusstageone, with English teacher being ReAtt~\citep{jiang-etal-2022-retrieval} trained on NQ~\citep{kwiatkowski-etal-2019-natural}. \textit{ii) Multilingual QA Pre-training} (MLQA-PT): \oursusstageone is further pre-trained as in \S\ref{sec:stage2_pretrain} to obtain the unsupervised \oursus. \textit{iii) Fine-tuning}: We first fine-tune \oursus on NQ to obtain the zero-shot \ourszs, which is then trained on supervised data from \textsc{Xor-TyDi QA} to derive \ours. We use the same training objective $\mathcal{L}_{\text{e2e}}$ as in MLQA-PT.


\subsection{Main Results}
\inlinetitle{XOR-Retrieve.}
Table~\ref{tab:xor_retrieve_results} shows the results on the dev set of XOR-Retrieve. \ours, which exclusively employs question-answer pairs for training, demonstrates a substantial performance advantage over all baselines that rely on passage labels for contrastive learning. This advantage is particularly pronounced under unsupervised and zero-shot settings, where both variants, \oursus and \ourszs, achieve improvements of more than 10\% over state-of-the-art methods ($p<0.001$).\footnote{Paired Student's t-test~\citep{dror-etal-2018-hitchhikers}.} The \textbf{Dense Retrieval} variant (i.e., w/ Dense) consistently outperforms other competitive baselines and is comparable to LAPCA with only 73\% of the parameters. This highlights that our approach is versatile and can be applied to enhance various kinds of retrievers.

\inlinetitle{XOR-Full.}
Table~\ref{tab:xor_full_reader_results} reports the results of \ours on XOR-Full. Both \ours and the variant employing dense retrieval achieve superior performance when compared to a series of baseline models and the prior state-of-the-art CORA model in all tested languages, showcasing an average improvement of up to 7.8\% ($p<0.001$). Compared to methods that rely on machine translation to generate a substantially larger pool of multilingual training data from English datasets, \ours is comparable to Sentri but falls behind LAPCA.\footnote{A direct comparison with Sentri and LAPCA is not feasible since the Wikipedia pages they employed as knowledge sources are different from ours and~\citet{asai2021one}.} The most pronounced performance gaps are in Bengali and Korean, with the fewest two training samples available within XOR-Full. We believe it is the translated QA pairs used by Sentri and LAPCA that alleviate such discrepancies, and further improvements are expected when integrating such augmented data.

\begin{figure}[t]
    \centering
    \includegraphics[width=\linewidth]{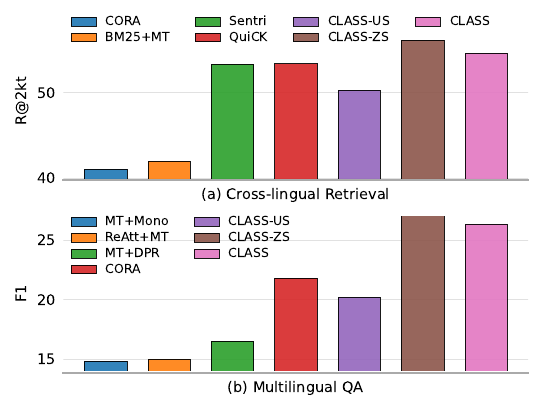}
    \caption{Zero-shot cross-lingual retrieval and multilingual QA results on unseen languages of MKQA. Macro average results across all test languages are reported. Languages included are: Da, De, Es, Fr, He, Hu, It, Km, Ms, Nl, No, Pl, Pt, Sv, Th, Tr, Vi, Zh-cn, Zh-hk, and Zh-tw.}
    \label{fig:mkqa_zs}
\end{figure}

\inlinetitle{MKQA.}

We assess the zero-shot performance of \ours in various unseen languages included in MKQA. Figure~\ref{fig:mkqa_zs} shows that in cross-lingual retrieval tasks, all variants of our method exhibit promising results. Notably, \oursus surpasses the supervised model CORA significantly, and further fine-tuning on English data leads to substantial improvements. Interestingly, \ours underperforms \ourszs, despite being further fine-tuned on multilingual data. Similar patterns are observed in the multilingual QA task, where \ourszs achieves the best zero-shot performance across unseen languages while supervised fine-tuning on XOR-Full hurts the generalisability. We attribute this phenomenon to three factors: 1) the limited number of queries in \textsc{Xor-TyDi QA} leads to overfitting to these specific languages, as we observe that the model's performance in both retrieval and QA tasks of MKQA decreases as the fine-tuning on \textsc{Xor-TyDi QA} continues; 2) the query topics differ, as MKQA was translated from NQ while \textsc{Xor-TyDi QA} questions were created by native speakers in target languages; 3) the answer type differs (free spans on \textsc{Xor-TyDi QA} \versus~WikiData aligned entities on MKQA). Our manual inspection reveals that \ours is more likely to generate free-span answers than \ourszs.
Detailed results in each language are in Appendix~\ref{appendix:detailed_zs_eval} Tables~\ref{tab:mkqa_retrieve_zs} and~\ref{tab:mkqa_qa_zs}.

\begin{table}[t]
\setlength{\tabcolsep}{2.8pt}
\footnotesize
\centering
\resizebox{\linewidth}{!}{\begin{tabular}{l|c|cccccccc}
    \toprule
    & Size & Ar & Bn & Fi & Ja & Ko & Ru & Te & Avg. \\
    \midrule
    \ourszs & 1.23B & 26.8 & 22.9 & 20.3 & 23.1 & 27.2 & 25.0 & 21.9 & 23.9 \\
    \midrule
    Gemma & 7B & 13.4 & 19.0 & 21.7 & 20.2 & 20.5 & 23.0 & 23.4 & 20.2 \\
    LLaMA3	& 8B & 22.7	& 13.2 & 22.9 & 17.8 & 19.0	& 19.2 & 28.9 & 20.5 \\
    \ours & 1.23B & 32.3 & 28.1 & 29.9	& 25.7 & 29.5 & 27.7 & 24.7 & 29.8 \\
    \bottomrule
\end{tabular}}
    \caption{F1 scores on XOR-Full under 5-shot learning settings. Gemma and LLaMA3 are RALM baselines. \ours is obtained through 5-shot fine-tuning over \ourszs.}
\label{tab:few_shot_results}
\end{table}
\subsection{Few-shot Results}
To further demonstrate the superiority of \ours under low-resource settings, we compare our method against retrieval-augmented language model (RALM) baselines, where all systems are provided 5-shot supervision.
The shots are sampled from the XOR-Full training data. 
For the RALM baselines we prompt 
a LLM (Llama3-8B~\citep{dubey2024llama3herdmodels} and Gemma-7B~\citep{gemmateam2024gemmaopenmodelsbased}) with the 5-shot instances and the retrieved passages to the query (retrieved by \ourszs). For \ours, we fine-tune \ourszs on the same five-shot examples.

As shown in Table~\ref{tab:few_shot_results}, \ours is significantly better than two RALM baselines (+9.3\%) despite having roughly 5$\times$ fewer parameters. Meanwhile, note that our zero-shot model \ourszs surpasses the two RALMs by 3.4\% (23.9\% \versus 20.5\%) without using any supervised data.
This demonstrates the superiority of \ours for zero and low-resource multilingual open-domain QA.

\subsection{Analysis}
We include quantitative and qualitative error analysis in Appendix~\ref{appendix:error_analysis} and additional numeric results in Appendix~\ref{appendix:more_analysis} (Figures~\ref{fig:results_against_tokens},~\ref{fig:retrieve_few_shot},~\ref{fig:num_of_infer_psgs})).

\begin{figure}[t]
    \centering
    \includegraphics[width=\linewidth]{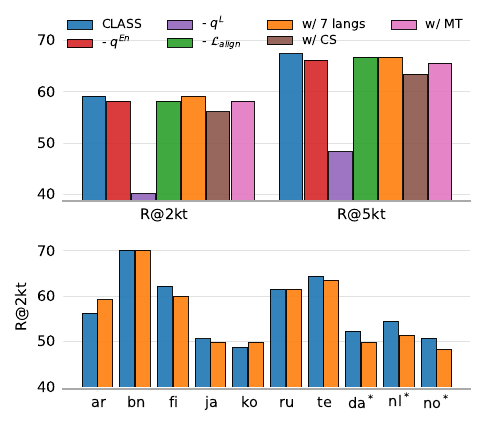}
    \caption{Ablations on cross-lingual retrieval pre-training, with results on the XOR-Retrieve dev set reported. $^{\ast}$ indicates unseen languages from MKQA.}
    \label{fig:stage1_ablations}
\end{figure}

\inlinetitle{Cross-lingual Retrieval Pre-training Ablations.}
We conduct ablation studies to understand the impact of different components in cross-lingual retrieval pre-training, with results shown in Figure~\ref{fig:stage1_ablations}.

\textbf{Effects of Learning from Parallel Queries.}\ \ Removing queries either in English (-$q^{En}$) or in target languages (-$q^{L}$) leads to performance degradation. Meanwhile, the cross-lingual alignment regularisation (-$\mathcal{L}^{align}$) benefits the model by ensuring consistent predictions across languages.

\textbf{Comparison with Different Parallel Query Sources.}\ \ When comparing the approaches of gathering parallel queries, our method outperforms code-switching (w/ CS), which creates pseudo-translations through lexicon replacement based on bilingual dictionaries, and machine translations (w/ MT). This inferiority is primarily attributed to the limited coverage of bilingual dictionaries and poor translation quality in low-resource languages.

\textbf{Sensitivity to Pre-training Language.}\ \ Removing the extra 8 high-resource languages (w/ 7 langs) does not impact average performance but \emph{affects specific low-resource languages} in \textsc{Xor-TyDi QA}. In particular, adding languages related to Telugu and Japanese (\eg Tamil \& Chinese) yields improvements. \emph{Moreover, including a wider range of languages improves generalisation to unseen low-resource languages with limited parallel Wikipedia links} (\eg adding German data enhances understanding of the West Germanic languages: Danish, Dutch, and Norwegian).

\begin{table}[t]
    \setlength{\tabcolsep}{2pt}
    \footnotesize
    \centering
    \resizebox{\linewidth}{!}{
    \begin{tabular}{l|cc|ccccc}
        \toprule
        \multirow{2}{*}[-1ex]{\bf Method} & \multicolumn{2}{c|}{\bf XOR-Retrieve} & \multicolumn{5}{c}{\bf XOR-Full} \\
        \cmidrule(lr){2-3} \cmidrule(lr){4-8}
        & \bf R{\scriptsize @}2kt & \bf R{\scriptsize @}5kt & \bf F1 & \bf EM & \bf BLEU & \bf R$^\text{L}${\scriptsize @}N & \bf R$^\text{M}${\scriptsize @}N \\
        \midrule
        \rowcolor{blue!15} \multicolumn{8}{c}{\bf\emph{Unsupervised}} \\
        \textbf{\oursus} (AB) & \bf 66.5 & \bf 73.1 & \bf 18.4 & \bf 12.0 & \bf 14.6 & 60.0 & 69.1 \\
        \ \ - MLQA-PT (A) & 59.1 & 67.4 & 5.7 & 3.9 & 4.0 & 55.7 & \bf 74.7 \\
        \ \ - Query TF (AC) & 66.1 & \bf 73.1 & 7.2 & 4.8 & 4.9 & \bf 60.1 & 65.4 \\
        \rowcolor{blue!15} \multicolumn{8}{c}{\bf\emph{Zero-shot}} \\
        \textbf{\ourszs} (ABD) & \bf 68.2 & \bf 75.2 & \bf 23.9 & \bf 15.8 & \bf 19.4 & \bf 59.2 & 69.1 \\
        \ \ - MLQA-PT (AD) & 62.9 & 71.1 & 13.7 & 8.1 & 8.3 & 57.0 & \bf 76.2 \\
        \ \ - Pre-train (D) & 27.6 & 36.3 & 15.4 & 9.6 & 11.0 & 52.5 & 58.6 \\
        \rowcolor{blue!15} \multicolumn{8}{c}{\bf\emph{Supervised}} \\
        \textbf{\ours} (ABDE) & \bf 71.6 & \bf 78.2 & 42.4 & 32.7 & 29.2 & 62.8 & \bf 78.4 \\
        \ \ - MLQA-PT (ADE) & 69.6 & 75.7 & \bf 42.5 & \bf 33.1 & 29.1 & \bf 63.1 & 77.8 \\
        \ \ - Pre-train (DE) &  62.8 & 69.3 & 41.9 & 32.6 & 28.7 & 62.4 & 71.7 \\
        \bottomrule
    \end{tabular}}
    \caption{Effects of two-stage pre-training. Results on the dev sets are reported. Symbols within brackets are described in Figure~\ref{fig:training_recipe}. R$^\text{L}${\scriptsize @}N and R$^\text{M}${\scriptsize @}N means the percentage of the questions whose top-N (N=100) passages contain an answer string in the target or any language.}
    \label{tab:two_stage_pretraining_ablation}
\end{table}

\inlinetitle{Effects of Two-stage Pre-training.}

We evaluate the efficacy of our two-stage proposed pre-training framework. Table~\ref{tab:two_stage_pretraining_ablation} showcases the performance on both XOR-Retrieve and XOR-Full under unsupervised, zero-shot, and supervised settings. Integrating multilingual QA pre-training dramatically boosts performance in both unsupervised and zero-shot scenarios. Merely employing cloze-style questions instead of transformed natural questions has minimal impacts on retrieval but yields sub-optimal QA results, highlighting the importance of synthetic natural questions in QA tasks. When discarding the entire pre-training process, we observe a notable drop in both datasets. In supervised settings, the advantages of pre-training diminish with labelled data. This is especially evident in XOR-Full, where the differences between \ours and the other two variants in QA and in-language retrieval (R$^L${\footnotesize @}N) results diminish. While pre-training significantly improves cross-lingual evidence retrieval (R$^M${\footnotesize @}N 71.7\% -> 78.4\%), \ours does not benefit from this, suggesting its heavy reliance on in-language evidence and inability to reason over cross-lingual evidence when generating answers. See Appendix~\ref{appendix:error_analysis} for more detailed error analysis.


\section{Related Work}

    
\inlinetitle{Multilingual Dense Retrieval.}
Dense retrievers adopt pre-trained language models and follow a dual-encoder architecture~\citep{karpukhin-etal-2020-dense} to encode queries and passages into dense vectors and calculate the similarity scores. Effective techniques were proposed to advance English dense retrievals, including hard negative mining~\citep{xiong2021approximate}, multi-vector representations~\citep{colbertv1}, and distilling from cross-encoder rerankers~\citep{ren-etal-2021-rocketqav2}. With the advent of multilingual pre-trained models, these techniques were adapted to improve cross-lingual dense retrievals~\citep{asai2021one, ren-etal-2022-empowering}. However, all these methods rely on passage labels for contrastive learning, which is challenging to obtain in cross-lingual settings. In contrast, our method explores a semi-supervised method and shows that a competitive cross-lingual retriever can be achieved using only query-answer pairs.


\inlinetitle{Multilingual Retrieval Pre-training.}
Large-scale unsupervised retrieval pre-training has significantly enhanced dense retrievers~\citep{gao-callan-2021-condenser, izacard2022unsupervised} in processing English texts. Pre-training has also been explored in cross-lingual and multilingual dense retrieval, with a particular emphasis on augmenting the cross-lingual alignment capabilities of models. LAPCA~\citep{lapca} is trained through extensive cross-lingual contrastive learning, employing texts from parallel Wikipedia pages and parallel texts generated by machine translation systems. DrDecr~\citep{li-etal-2022-learning-cross} learns from English models but operates on a smaller scale and relies on supervised parallel queries. In this work, we delve into the potential of large-scale unsupervised pre-training for cross-lingual dense retrieval and show that the resulting model exhibits high efficacy, outperforming many supervised ones.


\inlinetitle{Pre-training for Retrieval-Augmented Multilingual QA.}
In the context of English, jointly training a retriever and reader on supervised query-answer pairs~\citep{sachan2021endtoend, rag} or large-scale unsupervised data derived from masked salient span masking~\citep{realm, lee-etal-2022-need} have been shown to enhance the performance of both retrieval and question answering tasks. However, the application of such a joint training paradigm, whether in supervised training or unsupervised pre-training, has not been explored in cross-lingual and multilingual settings. Our study represents the first investigation into this issue and proposes a curated pre-training framework within a unified model to address both retrieval and question-answering tasks. We introduce a two-stage pre-training procedure to initially equip a multilingual model with robust cross-lingual retrieval abilities by learning from English experts and then gradually evolving it through exposure to large-scale multilingual QA pairs. This approach yields remarkable unsupervised results and significant performance improvements across unseen languages without annotated training data.

\section{Conclusion}
In this paper, we explore the potential of a unified model for both cross-lingual retrieval and multilingual QA tasks. By incorporating our proposed pre-training paradigm, \ours, the model's performance can be significantly improved, achieving both boosted retrieval and QA performance, while exhibiting impressive zero-shot transfer abilities to numerous unseen languages. Detailed ablations and thorough analyses are conducted to assess the efficacy of each component within our approach. Our future work aims at scaling \ours to a broader range of languages to further enhance the model's cross-lingual transfer performance.

\section*{Limitations}
The proposed pre-training framework incurs additional training costs when compared to standard supervised training, such as various pre-training data generation pipelines. The entire training pipeline requires approximately two weeks to complete with a maximum of 32 A100 GPUs. This could be less practical for researchers who do not have access to sufficient GPU resources. Nonetheless, common techniques such as \emph{gradient accumulation} can be applied to adapt our approach for training in a more academic setting, although more training time is required to achieve comparable results.

Both stages in our pre-training paradigm depend on the availability of parallel Wikipedia pages. This can pose a challenge when dealing with languages that have limited resources even in terms of monolingual texts. Our approach may fail when no language links exist between English and a specific low-resource language. One may resort to employing a multi-hop approach to discover parallel Wikipedia pages, by first searching for the language linked to the low-resource language within Wikipedia and then repeating this process iteratively until reaching the corresponding English page. Another option could be relying on the generalisation of the multilingual model by training it in closely-related languages. Our analysis has revealed that incorporating a high-resource language in the pre-training phase consistently results in improvements for other languages within the same language family (Figure~\ref{fig:stage1_ablations}), which makes this issue less of a concern. Nevertheless, it remains imperative to explore methods for reducing the reliance on parallel Wikipedia texts, as this is essential to scale our method to more diverse and unique languages, which is worth exploring as a future work.

This work does not examine the benefits of pre-training in a broader range of languages and the scaling effects of both model size and data size for multilingual QA tasks, which is an interesting research topic that should be addressed rigorously in the future.

As this work uses large language models for \emph{query transformation}, it is possible that undesirable biases (e.g., gender and cultural) inherent in these language models may be propagated to downstream systems. Furthermore, the extensive corpus of Wikipedia texts, drawn from a multitude of languages, could potentially introduce a diverse array of biases related to races and cultures to the pre-trained model. Assessing the magnitude of bias within the pre-training data and its subsequent impact on the model is an inherently intricate problem, which remains an open question for future research. Theoretically, our model can incorporate information extracted from any external corpus to generate answers to asked questions. This capability carries the potential for significant information leakage or the exposure of potentially toxic content from the corpus, which underscores the need for exercising caution when applying our method in sensitive domains.

\section*{Acknowledgement}
We thank the anonymous reviewers for their helpful feedback and suggestions. The first author is supported by the Graduate Research Scholarships funded by the University of Melbourne. This work was funded by the Australian Research Council, Discovery grant DP230102775. This research was undertaken using the LIEF HPCGPGPU Facility hosted at the University of Melbourne, which was established with the assistance of LIEF Grant LE170100200.

\bibliography{anthology,custom}

\clearpage
\section*{Overview of Appendix}
Our supplementary includes the following sections:
\begin{itemize}
    \item Section~\ref{appendix:exp_settings}: Experimental Settings, including implementation details, datasets, and compared baselines.
    \item Section~\ref{appendix:detailed_zs_eval}: Full zero-shot evaluation results on MKQA.
    \item Section~\ref{appendix:error_analysis}: Error analysis on multilingual open-domain question answering with quantitative and qualitative results.
    \item Section~\ref{appendix:more_analysis}: Additional numeric analysis.
    \item Section~\ref{appendix:icl_examples}: Prompts and examples for query transformation in each target language.
\end{itemize}

\appendix
\section{Experimental Settings}\label{appendix:exp_settings}
\subsection{Implementation Details}
\subsubsection{Parallel Queries Mining}\label{appendix:parallel_mining}
\begin{figure}[t]
    \centering
    \setlength{\abovecaptionskip}{-0.1cm}
    \setlength{\belowcaptionskip}{-0.3cm}
    \includegraphics[width=0.48
    \textwidth]{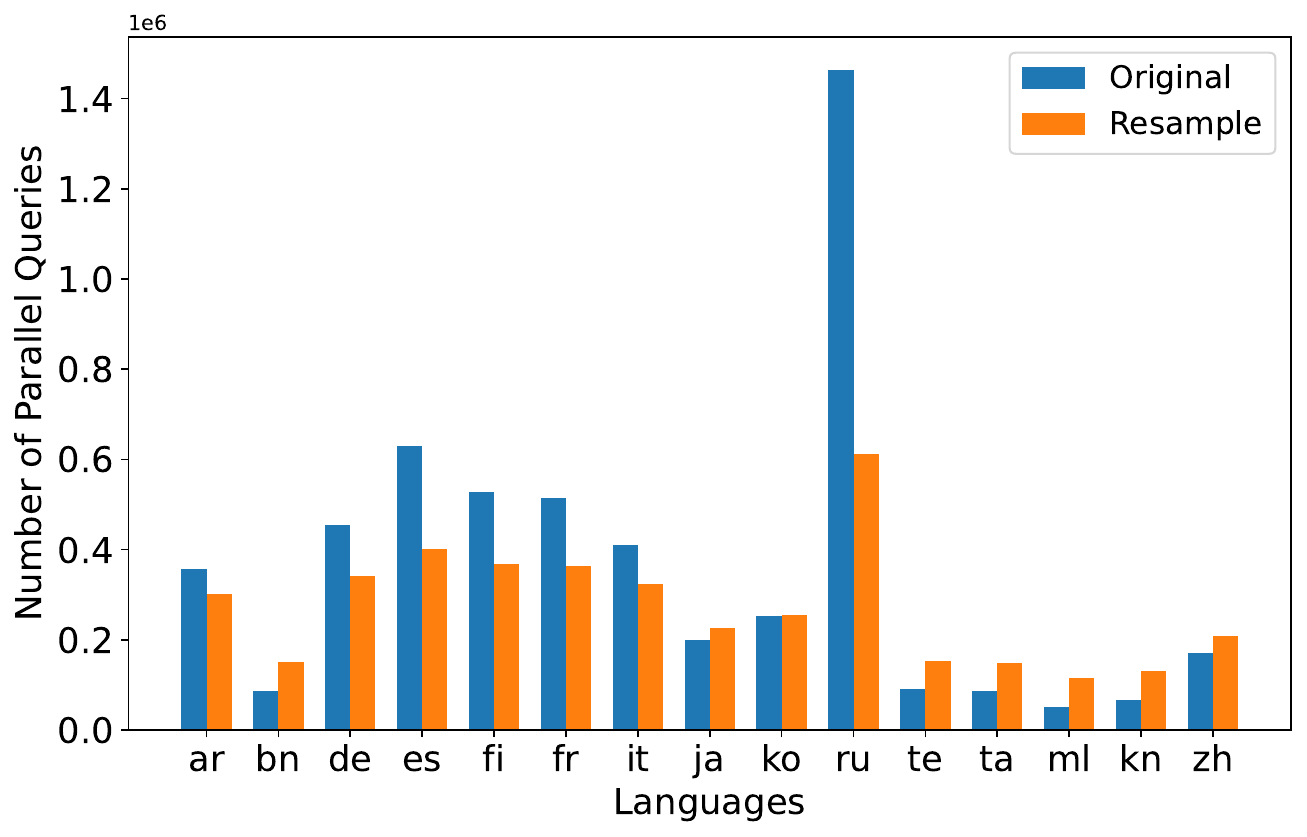}
    \caption{The number of mined parallel queries for each language pair \texttt{en-x}.}
    \label{fig:parallel_query_size}
\end{figure}

Our implementation encompasses 15 distinct languages, namely \textbf{Arabic}, \textbf{Bengali}, German, Spanish, \textbf{Finnish}, French, Italian, \textbf{Japanese}, \textbf{Korean}, \textbf{Russian}, \textbf{Telugu}, Tamil, Malayalam, Kannada, Chinese. Parallel queries are collected from parallel Wikipedia pages for each \texttt{en-x}. Using unsupervised contrastive learning, we adopt the approach in \citet{wang-etal-2022-english} to first pre-train a multilingual model \texttt{XLM-R}\footnote{\url{https://huggingface.co/xlm-roberta-large}} on English Wikipedia texts by taking the dropout as a form of data augmentation. The resulting model is proficient in generating universal cross-lingual sentence embeddings without the need for parallel data, demonstrating robust zero-shot cross-lingual transfer capabilities. Subsequently, we deploy the pre-trained model for extracting multilingual sentence embeddings and mining parallel queries for each \texttt{en-x} language pair. Empirically, we set the margin-score threshold to $1.5$ for most languages; however, for Japanese and Chinese, we observe improved performance with a larger threshold of $1.65$. This process yields 5.4 million examples for the training, with the number of parallel queries for each language pair \texttt{en-x} shown in Figure~\ref{fig:parallel_query_size}.

We employ a balanced sampling strategy to avoid the training bias towards high-resource languages. For $N$ number of languages $\{D_i\}_{i=1}^N$ with probabilities, $\{p_i\}_{i=1}^N$, we define the following multinomial distribution to sample from:
\begin{gather}
    p_i=\frac{f_i^\alpha}{\sum_{j=1}^Nf_j^\alpha}, \text{where}\ f_i = \frac{n_i}{\sum_{j=1}^Nn_j}, \nonumber
\end{gather}
where $\alpha$ is the sampling factor, which is set to $0.5$ by following~\citet{NEURIPS2019_c04c19c2} and $n_i$ is the total number of parallel queries in the $i$-th language. During training, we use this to determine $n'_i$, the number of parallel queries in each language; and top-$n'_i$ queries are used for training according to the margin-based scores. For every pair of mined query, we employ a state-of-the-art Named Entity Tagger from Stanza~\citep{qi-etal-2020-stanza}\footnote{\url{https://github.com/stanfordnlp/stanza}} to find salient entities within the English query and take all identified entities as answer candidates to construct cloze-style queries.

\subsubsection{Query Transformation}\label{appendix:query_transform}
We use ChatGPT to generate 32 meta-examples. We then employ LLaMA-2-7B\footnote{\url{https://huggingface.co/meta-llama/Llama-2-7b}} for query transformation by randomly sampling 3 meta-examples to construct prompts for each test instance, with the format as shown in Prompts~\ref{prompt:finnish_icl}, \ref{prompt:russian_icl}, \ref{prompt:japanese_icl}, \ref{prompt:korean_icl}, \ref{prompt:arabic_icl}, \ref{prompt:bengali_icl}, and \ref{prompt:telugu_icl}. We use Bloomz-7B\footnote{\url{https://huggingface.co/bigscience/bloomz-7b1}} for Telugu as we find LLaMA-2-7B does not work well in this language. The Question word \texttt{wh\_word} is chosen based on the entity type of the answer according to the heuristic rules in Table~\ref{tab:wh_word_heuristics}. Ultimately, 146K examples are generated per language, resulting in a total of 1M training instances.

\subsubsection{Training Details}\label{appendix:train_details}
We use \texttt{mt5-large}\footnote{\url{https://huggingface.co/google/mt5-large}} to initialise the model. In stage-1, we train the model for 64k steps on 32 A100 GPUs, which takes about one week to complete. The passages for all training queries are retrieved by the English teacher at once before training.
In stage-2, we further train the model for 16k steps on 16 A100 GPUs with roughly 4 days. We periodically update the retrieved passages for each training instance every 1k steps using the most recent model. 
For fine-tuning, we first train the model on NQ with 8k steps and fine-tune the model on XOR-Retrieve for 6k steps and 12k steps on XOR-Full, which takes about 19 hours and 156 hours to complete, respectively. Likewise, we also do passage refreshing periodically every 1k steps.

For all training stages, we use the same batch size of 64 queries with each paired with 100 retrieved passages and learning rate $5\times 10^{-5}$. We set $\alpha$ to 8 in all training loss functions. We set the maximum query and passage lengths to 50 and 200 for both training and evaluation.

For the \textbf{Dense Retrieval} variant, we follow the same training and hyperparameter settings. The only difference is that this configuration is significantly more efficient, with training time reduced by half for multilingual QA pre-training and fine-tuning.

\subsection{Datasets}
We used the following datasets for model evaluation in our experiments:
\begin{itemize}
    \item \boldtitle{XOR-Retrieve}\citep{asai-etal-2021-xor}. It is under the MIT License. It contains 15250 QA pairs for training and takes the 20190201 English Wikipedia dump which contains 18M passages as the retrieval database.
    \item \boldtitle{XOR-Full}\citep{asai-etal-2021-xor}. It is under the MIT License, containing 61360 training examples and a set of 43M passages as the retrieval corpus, collected from 20190201 Wikipedia dumps across 13 languages, namely English, Arabic, Finnish, Japanese, Korean, Russian, Bengali, Telugu, Indonesian, Thai, Hebrew, Swedish, and Spanish.
    \item \boldtitle{Natural Questions}\citep{kwiatkowski-etal-2019-natural}. It is under the Apache License and contains 79168 QA pairs.
    \item \boldtitle{MKQA}\citep{longpre-etal-2021-mkqa}. It is under the Apache License. This dataset covers 26 linguistically diverse languages, namely Arabic, Danish, German, English, Spanish, Finnish, French, Hebrew, Hungarian, Italian, Japanese, Korean, Khmer, Malay, Dutch, Norwegian, Polish, Portuguese, Russian, Swedish, Thai, Turkish, Vietnamese, Chinese (Simplified), Chinese (Hong Kong), and Chinese (Traditional). For the cross-lingual retrieval task, each language contains 6620 questions and the retrieval database consists of 18M English Wikipedia passages. For the multilingual QA task, each language contains 6758 questions and it uses the same retrieval database as XOR-Full. 
\end{itemize}

\subsection{Baselines}
\subsubsection{Cross-lingual Passage Retrieval}
We compare our proposed model with a range of strong baselines:
\begin{itemize}
    \item \boldtitle{mDPR.}This is the multilingual version of Dense Passage Retrieval (DPR)~\citep{karpukhin-etal-2020-dense} encoder, which undergoes initial training on English NQ queries followed by fine-tuning on XOR-Retrieve.
    \item \boldtitle{DPR+MT}\citep{asai-etal-2021-xor}. This is a translate-test baseline that involves the translation of queries into English during test time, followed by monolingual passage retrieval using the English DPR encoder.
    \item \boldtitle{CORA}\citep{asai2021one}. This method trains a multilingual DPR encoder iteratively, with positive and negative passages identified by a multilingual QA model.
    \item \boldtitle{Sentri}\citep{sorokin-etal-2022-ask}. An iterative self-training method that uses the latest retriever to identify positive and negative passages through answer string matching for updating the training dataset. Machine translation is used for data augmentation.
    \item \boldtitle{QuiCK}\citep{ren-etal-2022-empowering}. A knowledge distillation method that trains a multilingual bi-encoder retriever, learning from a query generator as the teacher. The query generator is also used for generating synthetic multilingual queries to enhance knowledge distillation.
    \item \boldtitle{DrDecr}\citep{li-etal-2022-learning-cross}. A multilingual ColBERT model that learns from an English ColBERT on parallel queries, sourced from both parallel corpora and human-translated gold queries released by XOR-Retrieve.
    \item \boldtitle{LAPCA}\citep{lapca}. A pre-training method that takes the first paragraphs of parallel Wikipedia pages as the parallel corpus for cross-lingual pre-training, with augmented data through machine translation.
    \item \boldtitle{SWIM-X}\citep{n2023leveraging}. A method that uses large language models to generate synthetic queries from unlabelled corpus with textual summary generation as an intermediate step. A multilingual dense retrieval model is fine-tuned exclusively on synthetic data.
\end{itemize}

\subsubsection{Multilingual Open Domain Question Answering}
\begin{itemize}
    \item \boldtitle{MT+DPR}\citep{asai-etal-2021-xor}. This represents the translate-test baseline, in which queries are translated into English and the answers are identified within English passages retrieved by the DPR+MT retriever. The English answer is then translated back to the target language if necessary.
    \item \boldtitle{ReAtt+MT}\citep{jiang-etal-2022-retrieval}. This is the English teacher employed in the cross-lingual retrieval pre-training. We use a state-of-the-art machine translation model\footnote{\url{https://huggingface.co/facebook/m2m100_418M}} to translate the queries into English at test time. It always retrieves passages from English Wikipedia and generates answers in English. The generated answer is translated back to the target language.
    \item \boldtitle{GMT+GS}\citep{asai-etal-2021-xor}. This pipeline follows the same procedure as \textbf{MT+DPR} except that we employ Google Search for passage retrieval and Google Machine Translation services for query and answer translation.
    \item \boldtitle{Monolingual baseline (BM25)}\citep{asai-etal-2021-xor}. Instead of using a multilingual DPR or an English DPR model with query translation, this baseline always retrieves the passage from the target language and extracts the answer using a multilingual reader.
    \item \boldtitle{MT+Mono}\citep{asai-etal-2021-xor}. This is a combination of the \textbf{BM25} and \textbf{MT+DPR} baselines, which first does monolingual QA for the target language using the BM25 method and resorts to the \textbf{MT+DPR} baseline if no answer is found.
    \item \boldtitle{Fusion-in-Decoder.}This encompasses a family of multilingual retrieval-augmented generation models, which take the passages returned by a multilingual retriever as inputs to generate the answer in the target language. \textbf{CORA}~\citep{asai2021one}, \textbf{Sentri}~\citep{ren-etal-2022-empowering} and \textbf{LAPCA}~\citep{lapca} are included in this family by using the passages returned by their respective retrievers.
\end{itemize}

\begin{table*}[t]
\setlength{\belowcaptionskip}{-0.2cm}
\setlength{\tabcolsep}{3pt}
\footnotesize
\centering
\resizebox{\linewidth}{!}{\begin{tabular}{l|ccccccccccccccccccccc}
    \toprule
    \bf Method & Da & De & Es & Fr & He & Hu & It & Km & Ms & Nl & No & Pl & Pt & Sv & Th & Tr & Vi & cn & hk & tw & Avg \\
    \midrule
    \rowcolor{blue!15} \multicolumn{22}{c}{\bf\emph{Unsupervised}} \\
    \bf \oursus & 50.5 & 53.4 & 53.8 & 53.9 & 44.1 & 49.1 & 52.6 & 39.8 & 55.3 & 53.3 & 49.5 & 52.6 & 50.4 & 52.5 & 54.9 & 50.9 & 48.0 & 48.0 & 46.3 & 46.4 & 50.3 \\
    \rowcolor{blue!15} \multicolumn{22}{c}{\bf\emph{Zero-shot}} \\
    BM25+MT & 44.1 & 43.3 & 44.9 & 42.5 & 36.9 & 39.3 & 40.1 & 31.3 & 42.5 & 46.5 & 43.3 & 46.5 & 45.7 & 49.7 & 46.5 & 42.5 & 43.5 & 37.5 & 37.5 & 36.1 & 42.0 \\
    \bf \ourszs & \bf 59.3 & \bf 58.9 & \bf 59.4 & \bf 59.2 & \bf 50.1 & \bf 54.0 & \bf 58.7 & \bf 46.2 & \bf 59.6 & \bf 60.4 & \bf 58.5 & \bf 57.5 & \bf 58.0 & \bf 59.4 & \bf 58.0 & \bf 55.1 & \bf 54.1 & \bf 52.1 & \bf 51.5 & \bf 51.4 & \bf 56.1 \\
    \ \ - MLQA-PT & 58.0 & 57.6 & 57.7 & 58.0 & 47.3 & 51.8 & 57.2 & 44.4 & 58.0 & 59.3 & 57.1 & 56.1 & 56.2 & 57.7 & 56.4 & 53.6 & 52.3 & 50.6 & 49.8 & 49.1 & 54.4 \\
    \ \ - Pre-train & 50.9 & 50.5 & 49.9 & 50.0 & 32.5 & 41.9 & 49.6 & 32.9 & 49.9 & 52.3 & 50.2 & 46.6 & 49.3 & 51.5 & 44.2 & 44.7 & 41.3 & 37.8 & 37.7 & 37.1 & 45.0 \\
    \rowcolor{blue!15} \multicolumn{22}{c}{\bf\emph{Supervised}} \\
    CORA & 44.5 & 44.6 & 45.3 & 44.8 & 27.3 & 39.1 & 44.2 & 22.2 & 44.3 & 47.3 & 48.3 & 44.8 & 40.8 & 43.6 & 45.0 & 34.8 & 33.9 & 33.5 & 41.5 & 41.0 & 41.1 \\
    Sentri & 57.6 & 56.5 & 55.9 & 55.1 & 47.9 & 51.8 & 54.3 & \bf 43.9 & 56.0 & 56.3 & 56.5 & 55.8 & 54.8 & 56.9 & 55.3 & 53.0 & \bf 54.4 & 50.2 & \bf 50.7 & 49.4 & 53.3 \\
    QuiCK & \bf 58.3 & 56.4 & 55.2 & 55.5 & 44.7 & 52.4 & 52.3 & 42.0 & 56.9 & 57.5 & \bf 57.0 & 54.9 & 54.7 & \bf 58.0 & 55.7 & 53.9 & 54.9 & 50.4 & 49.3 & 48.9 & 53.4 \\
    \bf \ours & 57.4 & \bf 57.5 & \bf 58.0 & \bf 57.8 & \bf 48.5 & \bf 52.5 & \bf 57.1 & 43.4 & \bf 58.2 & 58.4 & 56.7 & \bf 56.0 & \bf 56.4 & 57.6 & \bf 57.2 & \bf 54.2 & 52.5 & 51.3 & 49.9 & \bf 50.2 & \bf 54.6 \\
    \ \ - MLQA-PT & 56.9 & 57.3 & 57.2 & 57.0 & 47.3 & 51.8 & 56.2 & 42.9 & 57.6 & \bf 58.7 & 56.0 & 55.3 & 55.5 & 56.8 & 56.1 & 53.3 & 51.5 & 5\bf 1.4 & 49.9 & 49.4 & 53.9 \\
    \ \ - Pre-train & 56.5 & 55.3 & 55.9 & 55.1 & 44.8 & 50.8 & 55.0 & 41.3 & 56.4 & 57.4 & 55.8 & 53.3 & 54.8 & 56.5 & 53.7 & 51.9 & 49.6 & 47.3 & 46.4 & 45.8 & 52.2 \\
    \bottomrule
    \end{tabular}}
    \caption{Zero-shot cross-lingual retrieval results (R@2kt) on the MKQA dataset. "cn": "Zh-cn" (Chinese, simplified). "hk": "Zh-hk" (Chinese, Hong Kong). "tw": "Zh-tw" (Chinese, traditional).}
\label{tab:mkqa_retrieve_zs}
\end{table*}

\begin{table*}[t]
\setlength{\belowcaptionskip}{-0.15cm}
\setlength{\tabcolsep}{3pt}
\footnotesize
\centering
\resizebox{\linewidth}{!}{\begin{tabular}{l|ccccccccccccccccccccc}
    \toprule
    \bf Method & Da & De & Es & Fr & He & Hu & It & Km & Ms & Nl & No & Pl & Pt & Sv & Th & Tr & Vi & cn & hk & tw & Avg \\
    \midrule
    \rowcolor{blue!15} \multicolumn{22}{c}{\bf\emph{Unsupervised}} \\
    \bf \oursus & 24.9 & 27.4 & 29.1 & 27.1 & 12.9 & 21.7 & 25.2 & 9.3 & 26.3 & 27.0 & 25.0 & 23.7 & 22.4 & 26.0 & 13.2 & 22.8 & 17.5 & 7.3 & 8.9 & 6.3 & 20.2 \\
    \rowcolor{blue!15} \multicolumn{22}{c}{\bf\emph{Zero-shot}} \\
    ReAtt+MT & 22.4 & 23.9 & 21.6 & 23.5 & \bf 24.2 & 6.3 & 13.7 & 3.2 & 12.7 & 22.1 & 21.5 & 11.2 & 18.6 & 17.3 & 7.2 & 6.3 & 24.0 & \bf 10.8 & 4.7 & 4.0 & 15.0 \\
    MT+DPR & 26.2 & 25.9 & 28.4 & 21.9 & 8.9 & 15.7 & 25.1 & 1.2 & 12.6 & 28.3 & 18.3 & 24.6 & 24.7 & 19.7 & 6.9 & 18.2 & 15.1 & 3.3 & 3.8 & 3.8 & 16.5 \\
    \bf \ourszs & \bf 37.6 & \bf 38.5 & \bf 40.2 & \bf 37.6 & 17.0 & \bf 29.1 & \bf 36.2 & \bf 16.2 & \bf 36.9 & \bf 38.6 & \bf 37.4 & \bf 34.4 & \bf 33.6 & \bf 38.6 & \bf 18.9 & \bf 30.9 & \bf 29.6 & 8.7 & \bf 13.8 & \bf 8.5 & \bf 29.1 \\
    \rowcolor{blue!15} \multicolumn{22}{c}{\bf\emph{Supervised}} \\
    MT+Mono & 19.3 & 21.6 & 21.3 & 21.9 & 8.9 & 16.5 & 20.9 & 1.2 & 12.6 & 21.5 & 17.4 & 24.6 & 19.9 & 20.0 & 8.3 & 16.6 & 15.1 & 4.9 & 3.8 & 5.1 & 14.8 \\
    CORA & 30.4 & 30.2 & 32.0 & 30.8 & \bf 15.8 & 18.4 & 29.0 & 5.8 & 27.8 & 32.1 & 29.2 & 25.6 & 28.4 & 30.9 & 8.5 & 22.2 & 20.9 & 5.2 & 6.7 & 5.4 & 21.8 \\
    \bf \ours & \bf 33.4 & \bf 35.4 & \bf 37.5 & \bf 35.7 & 12.3 & \bf 27.7 & \bf 35.3 & \bf 10.2 & \bf 34.6 & \bf 36.1 & \bf 34.3 & \bf 31.9 & \bf 32.8 & \bf 33.3 & \bf 17.6 & \bf 29.3 & \bf 25.1 & \bf 8.6 & \bf 10.2 & \bf 7.4 & \bf 26.4 \\
    \bottomrule
    \end{tabular}}
    \caption{Zero-shot multilingual question answering results (F1) on the MKQA dataset. "cn": "Zh-cn" (Chinese, simplified). "hk": "Zh-hk" (Chinese, Hong Kong). "tw": "Zh-tw" (Chinese, traditional).}
\label{tab:mkqa_qa_zs}
\end{table*}


\section{Detailed Zero-shot Evaluation}\label{appendix:detailed_zs_eval}
\paragraph{Cross-lingual Retrieval.}
Table~\ref{tab:mkqa_retrieve_zs} presents the detailed result comparisons in each of the 20 unseen languages covered by MKQA. Notably, \ourszs outperforms other baselines significantly on average and achieves the best results in nearly all languages except for Vietnamese. Comparing the three variants of our method, fine-tuning on supervised English data significantly enhances cross-lingual transfer abilities to every unseen language (\ie \oursus vs \ourszs). However, fine-tuning \ourszs on a limited number of supervised multilingual data with a restricted language set does not lead to improved generalization performance, as indicated by the result comparison in every language between \ourszs and \ours. Furthermore, a decrease in performance is also observed in both supervised and zero-shot settings when either multilingual QA pre-training or the entire pre-training procedures are omitted, highlighting the effectiveness of our pre-training approach in enhancing cross-lingual ability.

\paragraph{Multilingual QA.}
Table~\ref{tab:mkqa_qa_zs} presents the detailed multilingual QA results for each of the 20 unseen languages covered by MKQA. We observe similar patterns where \oursus surpasses a range of machine-translation-based methods and \ourszs outperforms the supervised CORA by a significant margin. Further fine-tuning \ourszs on a limited number of supervised multilingual data with a restricted language set hampers its generalizability, with a decline in performance across all examined languages.


\begin{table}[t]
\setlength{\belowcaptionskip}{-0.35cm}
\footnotesize
\centering
\begin{tabular}{l|ccc}
    \toprule
    Model & F1 & EM & BLEU \\
    \midrule
    \ours & \bf 30.4 & 21.0 & \bf 20.6 \\
    \ours w/o MLQA-PT & 30.2 & \bf 21.4 & 20.3 \\ 
    \ours w/o Pre-train & 29.7 & 20.9 & 19.8 \\
    \bottomrule
    \end{tabular}
    \caption{Multilingual QA results on queries requiring cross-lingual evidence retrieval.}
\label{tab:cl_qa_results}
\end{table}

\section{Error Analysis}\label{appendix:error_analysis}
We add additional error analysis regarding the issue identified in multilingual QA (\ie XOR-Full).

\begin{table*}[t]
\setlength{\belowcaptionskip}{-0.3cm}
    \begin{minipage}{.32\linewidth}
      \footnotesize
      \setlength{\tabcolsep}{2.5pt}
      \centering
      \resizebox{\linewidth}{!}{
        \begin{tabular}{l|ccc}
            \toprule
            Model & R@2kt & R@5kt & R@10kt \\
            \midrule
            \ours & \bf 61.0 & \bf 70.6 & \bf 75.6 \\
            \ours w/o MLQA-PT & 59.0 & 68.8 & 74.6 \\
            \ours w/o Pre-train & 50.6 & 59.3 & 65.4 \\
            \bottomrule
        \end{tabular}}
        \caption*{(a) English or target language answer is in top-n retrieved tokens.}
        \label{subtab: cl_query_retrieval_en_target}
    \end{minipage}
    \hspace{0.01\linewidth}
    \begin{minipage}{.32\linewidth}
      \footnotesize
      \setlength{\tabcolsep}{2.5pt}
      \centering
      \resizebox{\linewidth}{!}{
        \begin{tabular}{l|ccc}
            \toprule
            Model & R@2kt & R@5kt & R@10kt \\
            \midrule
            \ours & \bf 50.4 & \bf 63.1 & \bf 70.8 \\
            \ours w/o MLQA-PT & 46.3 & 59.2 & 67.8 \\
            \ours w/o Pre-train & 32.7 & 42.1 & 50.4 \\
            \bottomrule
        \end{tabular}}
        \caption*{(b) Only English answer is in top-n retrieved tokens.}
        \label{subtab: cl_query_retrieval_en}
    \end{minipage}
    \hspace{0.01\linewidth}
    \begin{minipage}{.32\linewidth}
      \footnotesize
      \setlength{\tabcolsep}{2.5pt}
      \centering
      \resizebox{\linewidth}{!}{
        \begin{tabular}{l|ccc}
            \toprule
            Model & R@2kt & R@5kt & R@10kt \\
            \midrule
            \ours & 41.8 & 47.3 & 50.8 \\
            \ours w/o MLQA-PT & \bf 42.7 & \bf 47.9 & \bf 51.2 \\
            \ours w/o Pre-train & 41.0 & 46.9 & 50.4 \\
            \bottomrule
        \end{tabular}}
        \caption*{(c) Only Target language answer is in top-n retrieved tokens.}
        \label{subtab: cl_query_retrieval_target}
    \end{minipage}
    \caption{Retrieval accuracy of queries requiring answers based on English evidence.}
\label{tab:cl_query_retrieval}
\end{table*}

\begin{table*}[t]
\footnotesize
\centering
\setlength{\belowcaptionskip}{-0.4cm}
\begin{tabular}{c|cc}
    \toprule
    & Contain English Ans & No English Ans \\
    \midrule
    Contain Target Ans & F1: 41.0/EM: 30.6/BLEU: 33.2 & F1: 37.7/EM: 28.3/BLEU: 32.5 \\
    No Target Ans & F1: 13.5/EM: 2.1/BLEU: 12.9 & F1: 10.1/EM: 1.0/BLEU: 6.8 \\
    \bottomrule
    \end{tabular}
    \caption{Multilingual QA results on queries requiring cross-lingual evidence retrieval, grouped by whether the gold-standard answer string in English or the target language appears within the top-n retrieved tokens.}
\label{tab:breakdown_cl_qa_results}
\end{table*}

\subsection{Quantitative Analysis}
Our focus is on analysing the behaviour of our model when handling cross-lingual queries in XOR-Full. These queries require answers based on English evidence~\citep{asai-etal-2021-xor}. Initially, we analyse the retrieval accuracy of our model by assessing whether the top-n retrieved tokens contain the answer string in English or the target language.

As shown in Table~\ref{tab:cl_query_retrieval}, our pre-training method shows significant improvements in finding correct English evidence for those queries requiring cross-lingual evidence retrieval (e.g., 50.4\% -> 70.8\%) while maintaining competitive performance (Table~\ref{tab:cl_query_retrieval}(c)) in finding in-language (\ie the question language) evidence if there exists. Nevertheless, we have observed that these advancements do not translate into enhancements in the subsequent QA task, wherein the model is supposed to produce an answer in the same language as the question with English supporting documents. Table~\ref{tab:cl_qa_results} shows that our complete \ours model fails to achieve additional benefits in QA tasks despite its outstanding performance in retrieving cross-lingual evidence.

\begin{figure}[t]
    \setlength{\abovecaptionskip}{-0.02cm}
    \setlength{\belowcaptionskip}{-0.3cm}
    \centering
    \subfigure[Answer strings in target language or English are in top-100 retrieved passage]{
        \setlength{\abovecaptionskip}{-0.05cm}
        \label{fig:ca_score_en_tgt_ans}
        \includegraphics[width=0.45\textwidth]{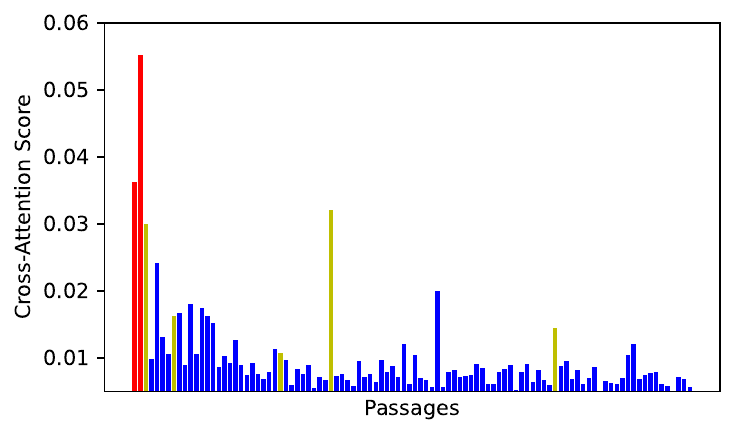}
    }
    \hspace{-0.02\textwidth}
    \subfigure[Only answer strings in English are in top-100 retrieved passages]{
        \setlength{\abovecaptionskip}{-0.05cm}
        \label{fig:ca_score_only_en_ans}
        \includegraphics[width=0.45\textwidth]{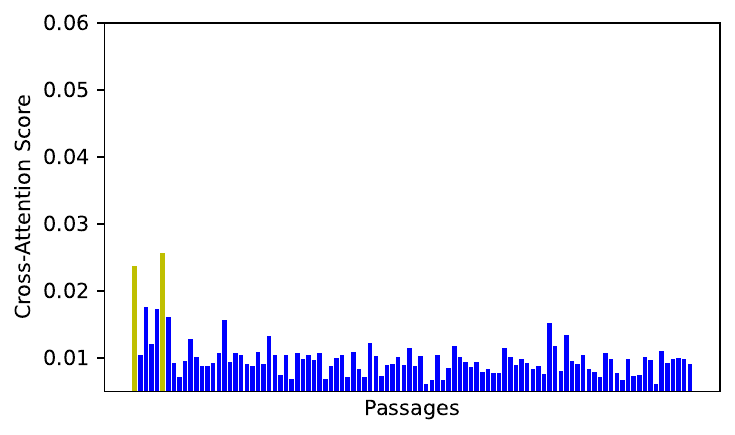}
    }
    \caption{Cross-Attention score to each of top-100 retrieved passages. Passages that contain the answer string in target languages or English are denoted with \textcolor{red}{red} and \textcolor{yellow}{yellow} bars, respectively.}
    \label{fig:ca_score}
\end{figure}

To gain deeper insights into the behaviour of our model, we specifically analyse its QA performance whenever the top-n retrieved evidence contains the gold answer in either English or the target language. As indicated in Table~\ref{tab:breakdown_cl_qa_results}, our model demonstrates reasonable performance only when the correct answer string is presented in the target language. However, it often fails to generate the correct answer when the gold standard answer is provided solely in English, despite our model being able to include the correct English answer in its top-10k retrieved tokens 71\% of the time. This indicates a deficiency in our model's ability to identify correct clues for QA among cross-lingual evidence. An example is shown in Figure~\ref{fig:ca_score}. In cases where the top 100 retrieved passages contain answer strings in the target language, our model tends to assign significantly higher scores to passages containing these target language answer strings. By contrast, when only English answer strings are present, the distribution of cross-attention scores across all retrieved passages becomes more uniform, leading to a general narrowing of the gap between positively relevant passages and irrelevant ones.

\begin{table*}[t]
    \centering
    \footnotesize
    \begin{tabular}{c}
        \toprule
        \includegraphics[width=0.97\linewidth]{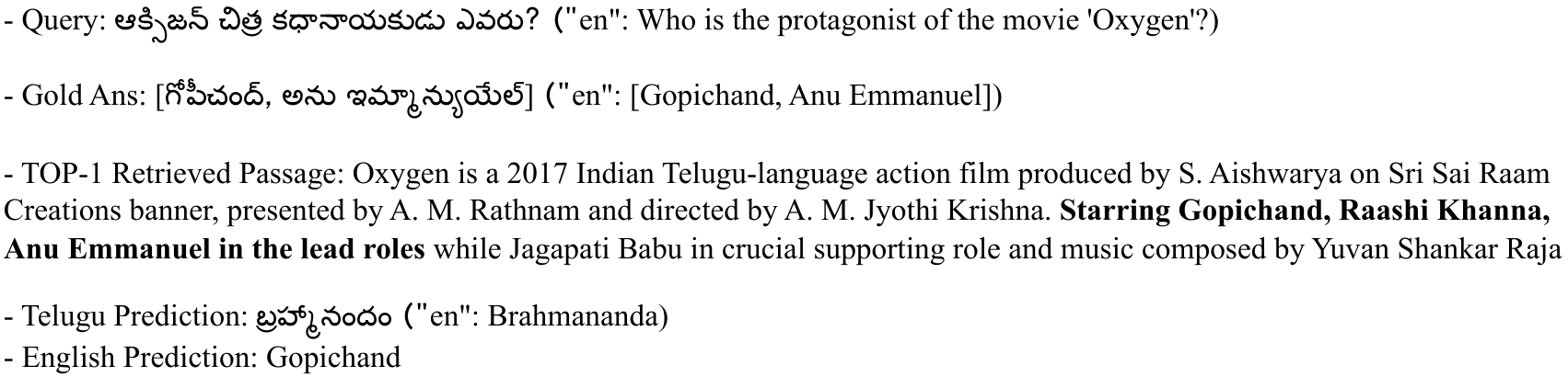} \\
        \bottomrule
    \end{tabular}
    \caption{An example of our model in finding correct evidence while failing to generate the right answer in the target language.}
    \label{tab:case_study}
\end{table*}

\begin{table*}[htb]
    \setlength{\belowcaptionskip}{-0.4cm}
    \setlength{\tabcolsep}{3pt}
    \footnotesize
    \centering
    \begin{tabular}{l|l|l}
        \toprule
        High Level Answer Category & Named Entity Types & Most appropriate \texttt{wh\_word} \\
        \midrule
        PERSON/NORP/ORG & PERSON, NORP, ORG & Who \\
        PLACE & GPE, LOC, FAC & Where \\
        THING & PRODUCT, EVENT, WORKOFART, LAW, LANGUAGE & What \\
        TEMPORAL & TIME, DATE & When \\
        NUMERIC & PERCENT, MONEY, QUANTITY, ORDINAL, CARDINAL & How much/How many \\
        \bottomrule
    \end{tabular}
    \caption{The heuristics rules for choosing the most appropriate question word based on named entity types (taken from~\citet{lewis-etal-2019-unsupervised}).}
    \label{tab:wh_word_heuristics}
\end{table*}

\subsection{Case Study}
As shown in Table~\ref{tab:case_study}, our model successfully retrieves the appropriate supporting document as its top-1 retrieval. However, it encounters challenges in generating Telugu answers, whereas it performs accurately in English. This highlights our model's inability to translate English evidence into answers in the target language, necessitating further efforts to enhance the model's capabilities in cross-lingual evidence reasoning and answer generation.

\section{More Analysis}\label{appendix:more_analysis}

\begin{figure}[t]
    \centering
    \setlength{\abovecaptionskip}{-0.05cm}
    \setlength{\belowcaptionskip}{-0.25cm}
    \includegraphics[width=0.452\textwidth]{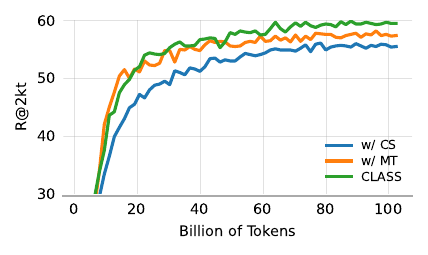}
    \caption{Performance evolution in stage-1 pre-training.}
    \label{fig:results_against_tokens}
\end{figure}
\paragraph{Performance Evolution during Pre-training.}
Figure~\ref{fig:results_against_tokens} illustrates the trajectory of the performance on the XOR-Retrieve cross-lingual retrieval task. As shown in the Figure, the use of code-switching consistently yields inferior results compared to \ours and the variant using machine translation. After training on around 45 billion tokens, \ours consistently outperforms MT, matching the performance of CS and MT with only 30\% and 50\% computation costs. This demonstrates greater training efficiency. The performance continues to improve over the next 50\% of the training tokens, implying that the scalability of pre-training data remains beneficial as training progresses.

\begin{figure}[t]
    \centering
    \setlength{\abovecaptionskip}{-0.05cm}
    \setlength{\belowcaptionskip}{-0.25cm}
    \includegraphics[width=0.452\textwidth]{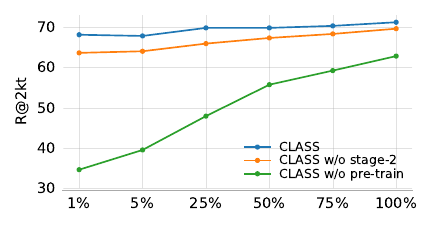}
    \caption{Scaling training data on cross-lingual retrieval.}
    \label{fig:retrieve_few_shot}
\end{figure}

\paragraph{Few-Shot Cross-lingual Retrieval.}
We consider a few-shot learning task with varying numbers of labelled training examples. Figure~\ref{fig:retrieve_few_shot} shows that \ours is consistently better than the other two variants, although the performance gap diminishes as more labelled data becomes available. Notably, as illustrated in Figure~\ref{fig:retrieve_few_shot}, the introduction of stage-2 pre-training results in a 75\% reduction in the required amount of labelled data. Furthermore, employing pre-training of both stages eliminates the need for any labelled data, in contrast to the approach that solely relies on supervised data for training (\ie \ours w/o pre-train).

\begin{figure}[t]
    \centering
    \setlength{\abovecaptionskip}{-0.05cm}
    \setlength{\belowcaptionskip}{-0.25cm}
    \includegraphics[width=0.452\textwidth]{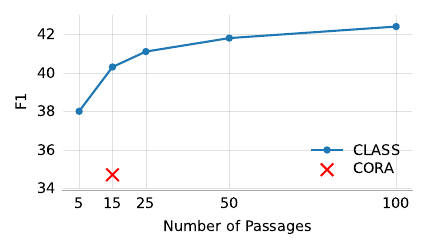}
    \caption{Effects of employing different numbers of retrieved passages for QA during inference time.}
    \label{fig:num_of_infer_psgs}
\end{figure}
\paragraph{Effects of Number of Retrieved Passages.}
Figure~\ref{fig:num_of_infer_psgs} reports the performance concerning the number of retrieved passages for QA during inference. We observe the performance improves consistently as the number of retrieved passages increases. \ours significantly outperforms CORA when using only top-5 retrieved passages, showcasing superior inference efficiency.

\begin{figure*}[t]
\centering
\setlength{\abovecaptionskip}{-0.02cm}
\begin{prompt}[title={Finnish Prompt}]
    You are an AI model that rewrites sentences into questions, using a given question word and answer. \\
    Rewrite this sentence "Strapping Young Lad (lyh. SYL) oli Devin Townsendin vuonna 1994 perustama kanadalainen metalliyhtye." into a natural question whose question word is "Milloin" and answer is "1994". Please respond in the format: "The transformed question is: \colorbox{lightblue}{Milloin Devin Townsend perusti kanadalaisen metalliyhtyeen Strapping Young Lad (lyh. SYL)?}"
\end{prompt}
\begin{prompt}[title={Russian Prompt}]
    You are an AI model that rewrites sentences into questions, using a given question word and answer.  \\
    Rewrite this sentence "\cyrins{В 215 году Цао Цао атаковал Чжан Лу и разгромил его в битве в проходе Янпингуань.}" into a natural question whose question word is "\cyrins{Кто}" and answer is "\cyrins{Чжан Лу}". Please respond in the format: "The transformed question is: \colorbox{lightblue}{\cyrins{Кто был атакован Цао Цао и разгромлен в битве в проходе Янпингуань в 215 году?}}"
\end{prompt}
\begin{prompt}[title={Japanese Prompt}]
    You are an AI model that rewrites sentences into questions, using a given question word and answer.  \\
    Rewrite this sentence "\begin{CJK}{UTF8}{min}熊野那智神社（くまのなちじんじゃ）は、宮城県名取市にある神社である。\end{CJK}" into a natural question whose question word is "\begin{CJK}{UTF8}{min}どこ\end{CJK}" and answer is "\begin{CJK}{UTF8}{min}宮城県\end{CJK}". Please respond in the format: "The transformed question is: \colorbox{lightblue}{\begin{CJK}{UTF8}{min}熊野那智神社はどこにある神社ですか？\end{CJK}}"
\end{prompt}
\begin{prompt}[title={Korean Prompt}]
    You are an AI model that rewrites sentences into questions, using a given question word and answer.  \\
    Rewrite this sentence "\begin{CJK}{UTF8}{mj}19세기 후반에 아일랜드에는 독립과 토지개혁을 요구하는 운동이 크게 확산되었다.\end{CJK}" into a natural question whose question word is "\begin{CJK}{UTF8}{mj}어디\end{CJK}" and answer is "\begin{CJK}{UTF8}{mj}아일랜드\end{CJK}". Please respond in the format: "The transformed question is: \colorbox{lightblue}{\begin{CJK}{UTF8}{mj}19세기 후반에 독립과 토지개혁을 요구하는 운동이 크게 확산된 나라는 어디입니까?\end{CJK}}"
\end{prompt} 
\begin{prompt}[title={Arabic Prompt}]
    You are an AI model that rewrites sentences into questions, using a given question word and answer.  \\
    Rewrite this sentence "\aratext{ولكن في مجال التعليم العالي تفتقر المسلمين، ونظرا لأفراد أسرهم وأقاربهم أخذها عن وظائف في دول الخليج وجنوب شرق آسيا (أساسا وسنغافورة، وألمانيا ماليزيا وبروناي في جنوب آسيا) نفسها في سن مبكرة.}" into a natural question whose question word is "\aratext{أين}" and answer is "\aratext{جنوب شرق آسيا}". Please respond in the format: "The transformed question is: \\ \colorbox{lightblue}{\aratext{أين يأخذ أفراد أسر المسلمين وأقاربهم وظائف غالبًا، مما يؤدي إلى فقدان الاهتمام بالتعليم العالي؟}}"
\end{prompt}
\begin{prompt}[title={Bengali Prompt}]
    You are an AI model that rewrites sentences into questions, using a given question word and answer.  \\
    Rewrite this sentence "{\bng bharte Hajar Hajar manuSh AnaHaer mara JaJ, ikn/tu dhr/mpRcarkra taedr pRit Udasiin.}" into a natural question whose question word is "{\bng ekathaJ}" and answer is "{\bng bhart}". Please respond in the format: "The transformed question is: \colorbox{lightblue}{{\bng ekathaJ Hajar Hajar manuSh AnaHaer mara JaJ EbNNG dhr/mpRcarkra taedr pRit Udasiin thaek?}}"
\end{prompt}
\begin{prompt}[title={Telugu Prompt}]
    \includegraphics[width=\textwidth]{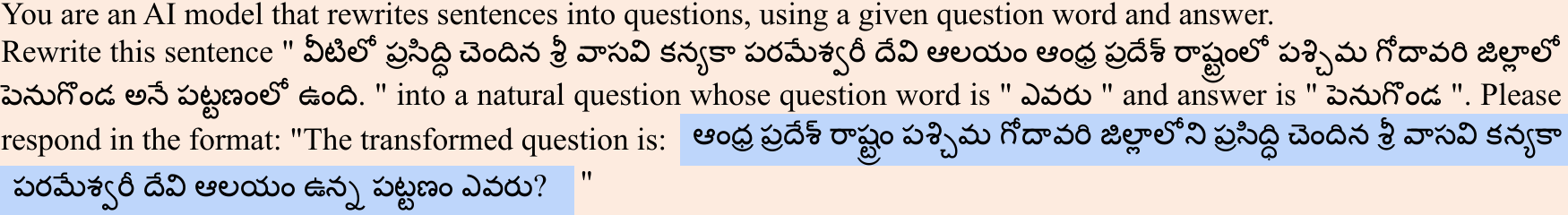}
\end{prompt}
\caption{Meta-examples obtained by prompting ChatGPT are shown for each language coverd by \textsc{Xor-TyDi QA}. \colorbox{lightblue}{Lightblue texts} indicate the transformed questions.}
\label{fig:chatgpt_prompt_example}
\end{figure*}
\section{Query Transformation Examples}\label{appendix:icl_examples}
Figure~\ref{fig:chatgpt_prompt_example} showcases examples illustrating the generation of meta-examples through prompting ChatGPT. Prompts~\ref{prompt:finnish_icl}, \ref{prompt:russian_icl}, \ref{prompt:japanese_icl}, \ref{prompt:korean_icl}, \ref{prompt:arabic_icl}, \ref{prompt:bengali_icl}, and \ref{prompt:telugu_icl} provide detailed illustrations of prompting a much smaller large language model, LLaMA-2-7B, to perform query transformation using In-Context Learning, which incorporates meta-examples in the target language $L$ from $\mathbb{K}$ into the prompt to guide the model's behaviour. The choice of the \emph{question word} is determined based on the detected entity type of the answer and the heuristic rules outlined in Table~\ref{tab:wh_word_heuristics}.

\begin{figure*}[t]
\begin{prompt}[title={Prompt \thetcbcounter: Finnish Example \& Translation}, label=prompt:finnish_icl]
    Rewrite sentences into short and precise questions, using given question words and answers:

    \ \ 

    Sentence: Toisaalta hän oli taiteiden suosija ja hänen valtakaudellaan Preussi sai haltuunsa suuren osan Puola-Liettuasta Puolan jaoissa vuosina 1793 ja 1795.
    
    Question word: Missä
    
    Answer: Preussi
    
    Transformed Question: Missä maassa taiteiden suosija hallitsi ja missä valtakunnassa saatiin haltuunsa suuri osa Puola-Liettuasta Puolan jaoissa vuosina 1793 ja 1795?

    \ \ 
    
    Sentence: Hän pelasi urallaan myös Ruotsissa ja Slovakiassa.
    
    Question word: Missä
    
    Answer: Slovakia
    
    Transformed Question: Missä maassa hän pelasi urallaan Ruotsin lisäksi?

    \ \ 
    
    Sentence: Barokin jälkeen concerto grossoja ovat säveltäneet muun muassa Heitor Villa-Lobos, Bohuslav Martinů, Alfred Schnittke ja Philip Glass.
    
    Question word: Kuka
    
    Answer: Bohuslav Martinů
    
    Transformed Question: Kuka säveltäjistä Heitor Villa-Lobosin, Alfred Schnittken ja Philip Glassin ohella on säveltänyt concerto grossoja barokin jälkeen?

    \ \ 

    Sentence: Hänen ajatteluunsa vaikuttivat muun muassa buddhalaiset ja taolaiset ideat, joihin hän tutustui Aasian matkoillaan, Mahatma Gandhin väkivallattomuusliike, sekä hänen katolinen uskontonsa. 
    
    Question word: Kuka
    
    Answer: Mahatma Gandhi
    
    Transformed Question: \colorbox{lightblue}{Kuka vaikutti hänen ajatteluunsa, mahtimaailmaan ja katoliseen uskontonsa?}
    \tcbline
    Rewrite sentences into short and precise questions, using given question words and answers:

    \ \ 

    Sentence: On the other hand, he/she was a fan of the arts and during his/her reign, Prussia took over a large part of Poland-Lithuania in the partitions of Poland in 1793 and 1795.
    
    Question word: Where
    
    Answer: Prussia
    
    Transformed Question: In which country did the lover of the arts rule and in which kingdom was a large part of Poland-Lithuania taken over during the partitions of Poland in 1793 and 1795?

    \ \ 
    
    Sentence: He/She also played in Sweden and Slovakia during her career.
    
    Question word: Where
    
    Answer: Slovakia
    
    Transformed Question: In which country did he/she play in his/her career besides Sweden?

    \ \ 
    
    Sentence: After the Baroque, concerto grossos have been composed by, among others, Heitor Villa-Lobos, Bohuslav Martinů, Alfred Schnittke and Philip Glass.
    
    Question word: Kuka
    
    Answer: Bohuslav Martinů
    
    Transformed Question: Besides Heitor Villa-Lobos, Alfred Schnittke and Philip Glass, which of the composers has composed concerto grossos after the Baroque?

    \ \ 

    Sentence: His/Her thinking was influenced, among other things, by Buddhist and Taoist ideas, which he/she got to know during his/her travels in Asia, Mahatma Gandhi's non-violence movement, and his/her Catholic religion.
    
    Question word: Who
    
    Answer: Mahatma Gandhi
    
    Transformed Question: \colorbox{lightblue}{Who influenced his/her thinking, the world of power and his/her Catholic religion?}
\end{prompt}
\end{figure*}

\begin{figure*}[t]
\begin{prompt}[title={Prompt \thetcbcounter: Russian Example \& Translation}, label=prompt:russian_icl]
    Rewrite sentences into short and precise questions, using given question words and answers: 

    \ \ 

    Sentence: \cyrins{Корабли проекта выполняли контроль за учениями ВМС стран НАТО в Норвежском и Средиземном морях, следили за корабельными и авианосными группами флотов США и Великобритании.}
    
    Question word: \cyrins{Кто}
    
    Answer: \cyrins{НАТО}
    
    Transformed Question: \cyrins{Кто выполнял контроль за учениями ВМС в Норвежском и Средиземном морях и следил за корабельными и авианосными группами флотов США и Великобритании?}

    \ \ 
    
    Sentence: \cyrins{1 апреля 1768 года Доверню назначают пенсию Королевской академии музыки в размере 1000 ливров как автору музыки.}
    
    Question word: \cyrins{Кто}
    
    Answer: \cyrins{Королевской академии музыки}
    
    Transformed Question: \cyrins{Кто 1 апреля 1768 года назначил пенсию в размере 1000 ливров Доверню как автору музыки?}

    \ \ 
    
    Sentence: \cyrins{Соф\'{и}я Шарл\'{о}тта Авг\'{у}ста (22 февраля 1847, Мюнхен — 4 мая 1897, Париж) — принцесса Баварская, герцогиня Баварская, позднее герцогиня Алансонская и Орлеанская.}
    
    Question word: \cyrins{Где}
    
    Answer: \cyrins{Мюнхен}
    
    Transformed Question: \cyrins{Где родилась София Шарлотта Августа, принцесса Баварская?}

    \ \ 

    Sentence: \cyrins{В первой половине XIX века паровозы в Россию, в основном, ввозились из-за рубежа.}
    
    Question word: \cyrins{Когда}
    
    Answer: \cyrins{XIX век}
    
    Transformed Question: \colorbox{lightblue}{\cyrins{Когда паровозы в Россию, в основном, ввозились из-за рубежа?}}
    \tcbline
    Rewrite sentences into short and precise questions, using given question words and answers: 

    \ \ 

    Sentence: The project's ships monitored NATO naval exercises in the Norwegian and Mediterranean Seas and monitored ship and aircraft carrier groups of the US and British navies.
    
    Question word: Who
    
    Answer: NATO
    
    Transformed Question: Who monitored naval exercises in the Norwegian and Mediterranean seas and monitored ship and aircraft carrier groups of the US and British fleets?

    \ \ 
    
    Sentence: On April 1, 1768, Dauvergne was awarded a pension from the Royal Academy of Music in the amount of 1000 livres as the author of music.
    
    Question word: Who
    
    Answer: Royal Academy of Music
    
    Transformed Question: Who, on April 1, 1768, awarded a pension of 1000 livres to Dovergne as the author of music?

    \ \ 
    
    Sentence: Sophia Charlotte Auguste (22 February 1847, Munich - 4 May 1897, Paris) - Princess of Bavaria, Duchess of Bavaria, later Duchess of Alençon and Orléans.
    
    Question word: Where
    
    Answer: Munich
    
    Transformed Question: Where was Sophia Charlotte Augusta, Princess of Bavaria born?

    \ \ 

    Sentence: In the first half of the 19th century, steam locomotives were mainly imported to Russia from abroad.
    
    Question word: When
    
    Answer: 19th century
    
    Transformed Question: \colorbox{lightblue}{When were steam locomotives mainly imported into Russia from abroad?}
\end{prompt}
\end{figure*}

\begin{figure*}[t]
\begin{prompt}[title={Prompt \thetcbcounter: Japanese Example \& Translation}, label=prompt:japanese_icl]
    Rewrite sentences into short and precise questions, using given question words and answers: 

    \ \ 

    \begin{CJK}{UTF8}{min}
    Sentence: 2月、竇憲は左校尉の耿夔を遣わし、金微山において北匈奴の単于を包囲しこれを大いに破り、単于の母の閼氏を捕虜とした。
    
    Question word: 誰
    
    Answer: 匈奴
    
    Transformed Question: 2月に金微山で竇憲の遣わした左校尉の耿夔が包囲し大いに破ったのは誰の単于ですか？

    \ \ 
    
    Sentence: この町を法人化する法はリチャード・キャズウェルが提出し、キャズウェルはここを本拠地とし、後の1776年から1780年までノースカロライナ州の初代知事となった。
    
    Question word: どこ
    
    Answer: ノースカロライナ州
    
    Transformed Question: リチャード・キャズウェルが初代知事となったのはどこですか？

    \ \ 
    
    Sentence: これより以前、司空張華は司馬倫に疎まれて誅殺されていた。
    
    Question word: 誰
    
    Answer: 張華
    
    Transformed Question: 誰がこれより以前に司馬倫に疎まれて誅殺されていたのですか？

    \ \ 

    Sentence: 魯迅はこの無支祁が孫悟空の先祖・源流ではないかと推測した。
    
    Question word: 誰
    
    Answer: 魯迅
    
    Transformed Question: \colorbox{lightblue}{誰はこの無支祁が孫悟空の先祖・源流ではないかと推測したのでしょうか？}
    \end{CJK}
    \tcbline
    Rewrite sentences into short and precise questions, using given question words and answers: 

    \ \ 

    Sentence: In February, Dou Xian sent Zuo's lieutenant, Geng Kui, to besiege and defeat the Northern Xiongnu Danyu at Jinweishan, and took Danyu's mother, the Yan family, prisoner.
    
    Question word: Who
    
    Answer: Xiongnu
    
    Transformed Question: In February, in Jinweishan, which was the land of Danyu that was besieged and severely defeated by Geng Ku, the commander of the left school sent by Dou Xian?

    \ \ 
    
    Sentence: The act to incorporate the town was introduced by Richard Caswell, who made it his home and later became North Carolina's first governor from 1776 to 1780.
    
    Question word: Where
    
    Answer: North Carolina
    
    Transformed Question: Where did Richard Caswell become the first governor?

    \ \ 
    
    Sentence: Before this, Zhang Hua was shunned by Sima Lun and killed.
    
    Question word: Who
    
    Answer: Zhang Hua
    
    Transformed Question: Who had been shunned and killed by Sima Lun before this?

    \ \ 

    Sentence: Lu Xun surmised that this Mujiqi was the ancestor and origin of Sun Wukong.
    
    Question word: Who
    
    Answer: Lu Xun
    
    Transformed Question: \colorbox{lightblue}{Who could have guessed that Mujiqi was the ancestor/origin of Son Goku?}
\end{prompt}
\end{figure*}

\begin{figure*}[t]
\begin{prompt}[title={Prompt \thetcbcounter: Korean Example \& Translation}, label=prompt:korean_icl]
    Rewrite sentences into short and precise questions, using given question words and answers: 

    \ \ 

    \begin{CJK}{UTF8}{mj}
    Sentence: 전투에서 승리한 뒤, 오버워치는 10년간 계속해서 평화를 지켰으나 내분으로 인해 해산되었다.
    
    Question word: 누구
    
    Answer: 오버워치
    
    Transformed Question: 누구가 전투에서 승리한 뒤 10년 동안 평화를 지키다가 내분으로 인해 해산되었나요?

    \ \ 
    
    Sentence: 그가 구단을 떠난 지 10년이 되는 2013년 4월, 스포르팅 리스본은 호날두를 100,000번째 회원으로 등록해 경의를 표했다.
    
    Question word: 누구
    
    Answer: 스포르팅 리스본
    
    Transformed Question: 누가 2013년 4월 그가 구단을 떠난 지 10년이 되는 해에 호날두를 100,000번째 회원으로 등록해 경의를 표했나요?

    \ \ 
    
    Sentence: 19세기 후반에 아일랜드에는 독립과 토지개혁을 요구하는 운동이 크게 확산되었다.
    
    Question word: 어디
    
    Answer: 아일랜드
    
    Transformed Question: 19세기 후반에 독립과 토지개혁을 요구하는 운동이 크게 확산된 나라는 어디입니까?

    \ \ 

    Sentence: 산탄젤로 다리 () 또는 하드리아누스의 다리는 로마에 있는 다리 가운데 하나이다.
    
    Question word: 어디
    
    Answer: 로마
    
    Transformed Question: \colorbox{lightblue}{산탄젤로 다리가 있는 곳은 어디인가?}
    \end{CJK}
    \tcbline
    Rewrite sentences into short and precise questions, using given question words and answers: 

    \ \ 

    Sentence: After winning the battle, Overwatch continued to maintain peace for 10 years, but was disbanded due to internal strife.
    
    Question word: Who
    
    Answer: Overwatch
    
    Transformed Question: Who won the battle, kept the peace for ten years, and then disbanded due to infighting?

    \ \ 
    
    Sentence: In April 2013, 10 years after he left the club, Sporting Lisbon paid tribute to Ronaldo by registering him as their 100,000th member.
    
    Question word: Who
    
    Answer: Sporting Lisbon
    
    Transformed Question: Who paid tribute to Ronaldo by registering him as their 100,000th member in April 2013, marking 10 years since he left the club?

    \ \ 
    
    Sentence: In the late 19th century, movements calling for independence and land reform spread widely in Ireland.
    
    Question word: Where
    
    Answer: Ireland
    
    Transformed Question: In which country did the movement calling for independence and land reform spread significantly in the late 19th century?

    \ \ 

    Sentence: Ponte Sant'Angelo () or Hadrian's Bridge is one of the bridges in Rome.
    
    Question word: Where
    
    Answer: Rome
    
    Transformed Question: \colorbox{lightblue}{Where is the Ponte Sant'Angelo?}
\end{prompt}
\end{figure*}

\begin{figure*}[t]
\begin{prompt}[title={Prompt \thetcbcounter: Arabic Example \& Translation}, label=prompt:arabic_icl]
    Rewrite sentences into short and precise questions, using given question words and answers: 

    \ \ 

    Sentence: \aratext{ولد كلاي في مقاطعة هانوفر بولاية فرجينيا في عام 1777، ولكنه انتقل إلى ليكسينغتون، كنتاكي في عام 1797.}
    
    Question word: \aratext{أين}
    
    Answer: \aratext{كنتاكي}
    
    Transformed Question: \aratext{أين انتقل ولد كلاي بعد ميلاده في مقاطعة هانوفر، ولاية فرجينيا في عام 1797؟}

    \ \ 
    
    Sentence: \aratext{وكان نظام شركة سي بي إس هو الأكثر تقدما وفاز بالمنافسة في كل مرة.}
    
    Question word: \aratext{من}
    
    Answer: \aratext{شركة سي بي إس}
    
    Transformed Question: \aratext{من كان لديه النظام الأكثر تقدما وفاز بالمنافسة في كل مرة؟ }

    \ \ 
    
    Sentence: \aratext{كما الخطوط الجوية البريطانية تشغل أيضا صالة نادي التنفيذي بين البوابات B21 و B23.}
    
    Question word: \aratext{من}
    
    Answer: \aratext{الخطوط الجوية البريطانية}
    
    Transformed Question: \aratext{من يشغل صالة نادي التنفيذي بين البوابات B21 و B23؟ }

    \ \ 

    Sentence: \aratext{وقد افتتح حديثاً في قصر العظم  قاعتين تضمان بانوراما لحرب تشرين لتحريرية عام 1973، وبانوراما لحرب تموز 2006.}
    
    Question word: \aratext{أين}
    
    Answer: \aratext{قصر العظم }
    
    Transformed Question: \colorbox{lightblue}{\aratext{أين بانوراما لحرب تشرين لتحريرية عام 1973؟}}
    \tcbline
    Rewrite sentences into short and precise questions, using given question words and answers: 

    \ \ 

    Sentence: Clay was born in Hanover County, Virginia in 1777, but moved to Lexington, Kentucky in 1797.
    
    Question word: Where
    
    Answer: Kentucky
    
    Transformed Question: Where did Clay move after his birth in Hanover County, Virginia in 1797?

    \ \ 
    
    Sentence: The CPS system was the most advanced and won the competition.
    
    Question word: Who
    
    Answer: CPS system
    
    Transformed Question: Who had the most advanced system and won the competition?

    \ \ 
    
    Sentence: British Airways also operates the Teen Club lounge between gates B21 and B23.
    
    Question word: Who
    
    Answer: British Airways
    
    Transformed Question: Who operates the Executive Club lounge between gates B21 and B23?

    \ \ 

    Sentence: Two halls were recently opened in Al-Azm Palace containing a panorama of the October Liberation War of 1973, and a panorama of the July War of 2006.
    
    Question word: Where
    
    Answer: Al-Azm Palace
    
    Transformed Question: \colorbox{lightblue}{Where is the panorama of the October Liberation War of 1973?}
\end{prompt}
\end{figure*}

\begin{figure*}[t]
\begin{prompt}[title={Prompt \thetcbcounter: Bengali Example \& Translation}, label=prompt:bengali_icl]
    Rewrite sentences into short and precise questions, using given question words and answers: 

    \ \ 

    Sentence: {\bng JatRapeth sbar Aaeg edRoupdii pRaN Haran.}
    
    Question word: {\bng ek}
    
    Answer: {\bng edRoupdii}
    
    Transformed Question: {\bng JatRapeth sbar Aaeg ek pRaN Haran?}

    \ \ 
    
    Sentence: {\bng Apridek katarer rajdhanii edaHaet raishJar EkiT s/thaJii duutabas rJeech.}
    
    Question word: {\bng ekathaJ}
    
    Answer: {\bng katar}
    
    Transformed Question: {\bng raishJar s/thaJii duutabasiT ekathaJ Abis/tht?}

    \ \ 
    
    Sentence: {\bng bharte Hajar Hajar manuSh AnaHaer mara JaJ, ikn/tu dhr/mpRcarkra taedr pRit Udasiin.}
    
    Question word: {\bng ekathaJ}
    
    Answer: {\bng bhart}
    
    Transformed Question: {\bng ekathaJ Hajar Hajar manuSh AnaHaer mara JaJ EbNNG dhr/mpRcarkra taedr pRit Udasiin thaek?}

    \ \ 

    Sentence: {\bng EiT {O}JaishNNGTn -Er isJaTl-E Abis/tht ekhala jaJgaJ EkiT maechr bajar. }
    
    Question word: {\bng ekathaJ}
    
    Answer: {\bng isJaTl}
    
    Transformed Question: \colorbox{lightblue}{\bng EiT {O}JaishNNGTn ekathaJ ekhala jaJgaJ EkiT maechr bajar?}
    \tcbline
    Rewrite sentences into short and precise questions, using given question words and answers: 

    \ \ 

    Sentence: Draupadi was the first to die on the journey.
    
    Question word: Who
    
    Answer: Draupadi
    
    Transformed Question: Who died first on the journey?

    \ \ 
    
    Sentence: In addition, Russia has a permanent embassy in Doha, the capital of Qatar.
    
    Question word: Where
    
    Answer: Qatar
    
    Transformed Question: Where is the permanent embassy of Russia located?

    \ \ 
    
    Sentence: Thousands of people die of starvation in India, but missionaries are indifferent to them.
    
    Question word: Where
    
    Answer: India
    
    Transformed Question: Where are thousands of people dying of starvation and the missionaries are indifferent to them?

    \ \ 

    Sentence: It is an open-air fish market located in Seattle, Washington.
    
    Question word: Where
    
    Answer: Seattle
    
    Transformed Question: \colorbox{lightblue}{Where is an open air fish market in Washington?}
\end{prompt}
\end{figure*}

\begin{figure*}[t]
\begin{prompt}[title={Prompt \thetcbcounter: Telugu Example \& Translation}, label=prompt:telugu_icl]
    \includegraphics[width=\textwidth]{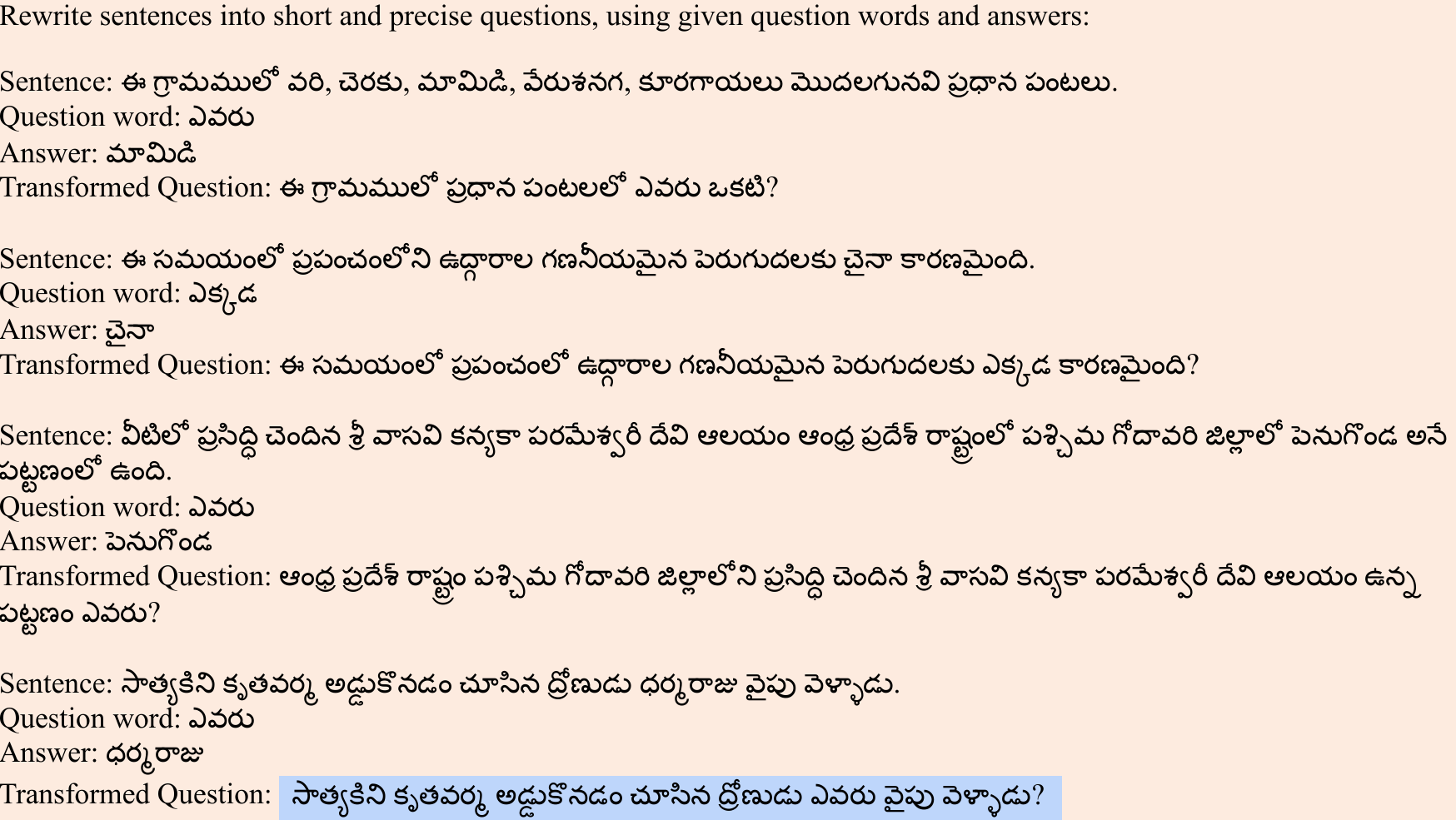}
    \tcbline
    Rewrite sentences into short and precise questions, using given question words and answers: 

    \ \ 

    Sentence: The main crops in this village are rice, sugarcane, mango, groundnut, vegetables etc.
    
    Question word: Who
    
    Answer: mango
    
    Transformed Question: Which is one of the main crops in this village?

    \ \ 
    
    Sentence: China accounted for a significant increase in world emissions during this period.
    
    Question word: Where
    
    Answer: China
    
    Transformed Question: Where in the world has caused the significant increase in emissions during this time?

    \ \ 
    
    Sentence: Among these, the famous Sri Vasavi Kanyaka Parameshwari Devi Temple is located in the town of Penugonda in the West Godavari district of the state of Andhra Pradesh.
    
    Question word: Who
    
    Answer: Penugonda
    
    Transformed Question: Which town in West Godavari district of Andhra Pradesh state has the famous Sri Vasavi Kanyaka Parameshwari Devi temple?

    \ \ 

    Sentence: Seeing Satyaki being stopped by Kritavarma, Drona went towards Dharmaraja.
    
    Question word: Who
    
    Answer: Dharmaraja
    
    Transformed Question: \colorbox{lightblue}{To whom did Drona go when he saw Kritavarma stopping Satyaki?}
\end{prompt}
\end{figure*}

\end{document}